\documentclass[review]{fcs}
\usepackage[utf8]{inputenc} % allow utf-8 input
\usepackage[T1]{fontenc}    % use 8-bit T1 fonts
\usepackage{microtype}
\usepackage{graphicx}
\usepackage{subfigure}
\usepackage{balance}
\usepackage{booktabs} % for professional tables
\usepackage{amsthm}
\usepackage{array}
\usepackage{graphicx}
\usepackage{clrscode}
\usepackage{subfigure}
\usepackage{multirow}
\usepackage{multicol}
\usepackage{float}
\usepackage{color}
\usepackage{xcolor}
\usepackage{amsopn}
\usepackage{mathrsfs}
\usepackage{mathtools}
\usepackage{amssymb}
\usepackage{amsmath}
\usepackage{booktabs}
\usepackage{arydshln}
\usepackage{hyperref}
\usepackage{blkarray}
\usepackage{enumerate}
\usepackage{courier}
\usepackage{mathrsfs}
\usepackage{rotating}
\usepackage{bm}
\usepackage{subfigure}
\usepackage{array}
\usepackage{ragged2e}
\usepackage{hyperref}
\usepackage{amsmath}
\usepackage{threeparttable} 
\usepackage{pifont}
\usepackage{comment}
\usepackage{epigraph}
\usepackage{url}

\theoremstyle{remark}
\newtheorem*{remark}{Remark}
\newcommand{\whitecircle}[1]{\ding{\numexpr171+#1\relax}}
\renewcommand{\thefootnote}{\fnsymbol{footnote}}

\title{A Survey on Large Language Model based Autonomous Agents}

\author{Lei Wang}
\author{Chen Ma\footnotemark[1]}
\author{Xueyang Feng\footnotemark[1]}
\author{Zeyu Zhang}
\author{Hao Yang}
\author{Jingsen Zhang}
\author{Zhi-Yuan Chen}
\author{Jiakai Tang}
\author*{Xu Chen}
\author*{Yankai Lin}
\author{Wayne Xin Zhao}
\author{Zhewei Wei}
\author{Ji-Rong Wen}

\address{Gaoling School of Artificial Intelligence, Renmin University of China, Beijing, 100872, China}

\fcssetup{
  received       = {month dd, yyyy},
  accepted       = {month dd, yyyy},
  corr-email     = {xu.chen@ruc.edu.cn;yankailin@ruc.edu.cn},
  doi={10.1007/s11704-024-40231-1}
}
\begin{abstract}
Autonomous agents have long been a research focus in academic and industry communities.
Previous research often focuses on training agents with limited knowledge within isolated environments, which diverges significantly from human learning processes, and makes the agents hard to achieve human-like decisions.
Recently, through the acquisition of vast amounts of web knowledge, large language models (LLMs) have shown potential in human-level intelligence, leading to a surge in research on LLM-based autonomous agents.
In this paper, we present a comprehensive survey of these studies, delivering a systematic review of LLM-based autonomous agents from a holistic perspective.
We first discuss the construction of LLM-based autonomous agents, proposing a unified framework that encompasses much of previous work. 
Then, we present an overview of the diverse applications of LLM-based autonomous agents in social science, natural science, and engineering.
Finally, we delve into the evaluation strategies commonly used for LLM-based autonomous agents.
Based on the previous studies, we also present several challenges and future directions in this field.
\end{abstract}

\begin{document}
\renewcommand{\thefootnote}{\fnsymbol{footnote}}
\footnotetext[1]{Both authors contribute equally to this paper.}
\renewcommand{\thefootnote}{\arabic{footnote}}

\begin{figure*}[t]
    \centering
    \setlength{\fboxrule}{0.pt}
    \setlength{\fboxsep}{0.pt}
    \fbox{
        \includegraphics[width=1.\linewidth]{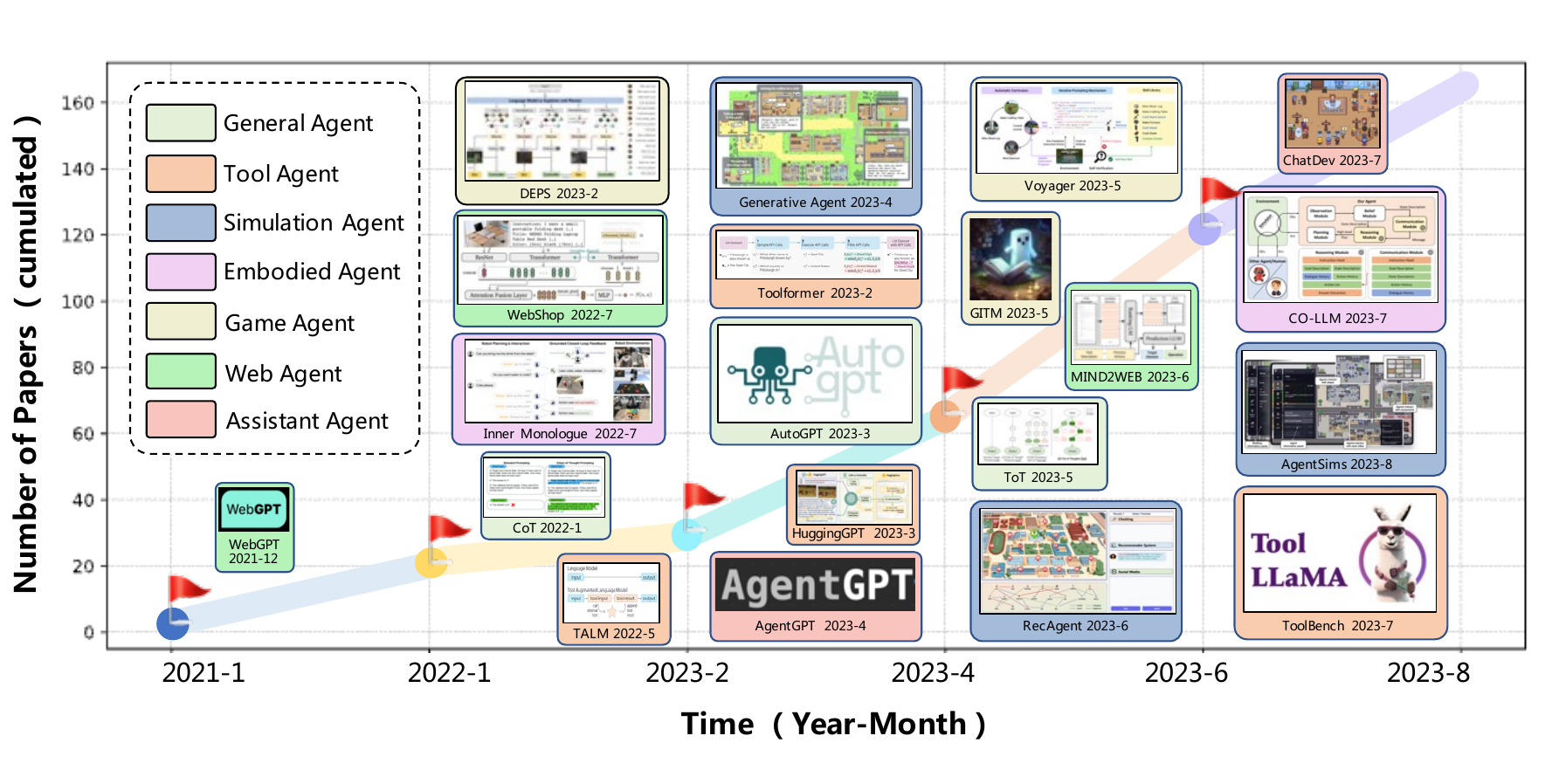}
    }
    % \vspace{-0.3cm}
    \caption{
    Illustration of the growth trend in the field of LLM-based autonomous agents.
    We present the cumulative number of papers published from January 2021 to August 2023.
    We assign different colors to represent various agent categories. For example, a game agent aims to simulate a game-player, while a tool agent mainly focuses on tool using. 
    For each time period, we provide a curated list of studies with diverse agent categories.
    }
    % \vspace{-0.3cm}
    \label{grow}
\end{figure*}

\section{Introduction}

\epigraph{\textit{“An autonomous agent is a system situated within and a part of an environment that senses that environment and acts on it, over time, in pursuit of its own agenda and so as to effect what it senses in the future.”}}{Franklin and Graesser (1997)}

Autonomous agents have long been recognized as a promising approach to achieving artificial general intelligence (AGI), which is expected to accomplish tasks through self-directed planning and actions.
In previous studies, the agents are assumed to act based on simple and heuristic policy functions, and learned in isolated and restricted environments~\cite{mnih2015human,lillicrap2015continuous,schulman2017proximal,haarnoja2018soft,brown2020language,radford2019language}. 
Such assumptions significantly differs from the human learning process, since the human mind is highly complex and individuals can learn from a much wider variety of environments.
Because of these gaps, the agents obtained from previous studies are usually far from replicating human-level decision processes, especially in unconstrained, open-domain settings.

In recent years, large language models (LLMs) have achieved notable successes, demonstrating significant potential to achieve human-like intelligence~\cite{achiam2023gpt,radford2019language,brown2020language,Claude2,touvron2023llama,touvron2023llama2}. This capability arises from leveraging comprehensive training datasets alongside a substantial number of model parameters.
Building upon this capability, there has been a growing research area that employs LLMs as central controllers to construct autonomous agents to obtain human-like decision-making capabilities~\cite{chen2019generative,shinn2023reflexion,shen2023hugginggpt,qin2023toolllm,schick2023toolformer,zhu2023ghost,sclar2023minding}.

Compared to reinforcement learning, LLM-based agents possess more comprehensive internal world knowledge, enabling them to perform informed actions even without training on specific domain data. Furthermore, LLM-based agents can offer natural language interfaces for human interaction, providing greater flexibility and enhanced explainability.

Along this direction, researchers have developed numerous promising models (see Figure~\ref{grow} for an overview), where the key idea is to equip LLMs with human capabilities such as memory and planning to make them behave like humans and complete various tasks effectively.
Previously, these models were proposed independently, with limited efforts made to summarize and compare them holistically.
However, we believe that a systematic summary of this rapidly developing field is of great significance for a comprehensive understanding of it and is beneficial in inspiring future research.

In this paper, we conduct a comprehensive survey of the field of LLM-based autonomous agents.
We organize our survey around three key aspects: construction, application, and evaluation of LLM-based autonomous agents.
For agent construction, we focus on two problems, that is, 
(1) how to design the agent architecture to better leverage LLMs, 
and (2) how to inspire and enhance the agent capability to complete different tasks.
Intuitively, the first problem aims to build the hardware fundamentals for the agent, while the second problem focuses on providing the agent with software resources.
For the first problem, we present a unified agent framework, which can encompass most of the previous studies.
For the second problem, we provide a summary on the commonly-used strategies for agents' capability acquisition.
In addition to discussing agent construction, we also provide a systematic overview of the applications of LLM-based autonomous agents in social science, natural science, and engineering.
Finally, we delve into the strategies for evaluating LLM-based autonomous agents, focusing on both subjective and objective strategies.

In summary, this survey conducts a systematic review and establishes comprehensive taxonomies for existing studies in the burgeoning field of LLM-based autonomous agents.
Our focus encompasses three primary areas: the construction of agents, their applications, and methods of evaluation.
Drawing from a wealth of previous studies, we identify various challenges in this field and discuss potential future directions.
% We believe that this field is still in its early stages; hence, we maintain a curated repository to keep track of ongoing studies at https://github.com/Paitesanshi/LLM-Agent-Survey.
We expect that our survey can provide newcomers of LLM-based autonomous agents with a comprehensive background knowledge, and also encourage further groundbreaking studies.

\begin{figure*}[t]
    \centering
    \setlength{\fboxrule}{0.pt}
    \setlength{\fboxsep}{0.pt}
    \fbox{
        \includegraphics[width=1.\linewidth]{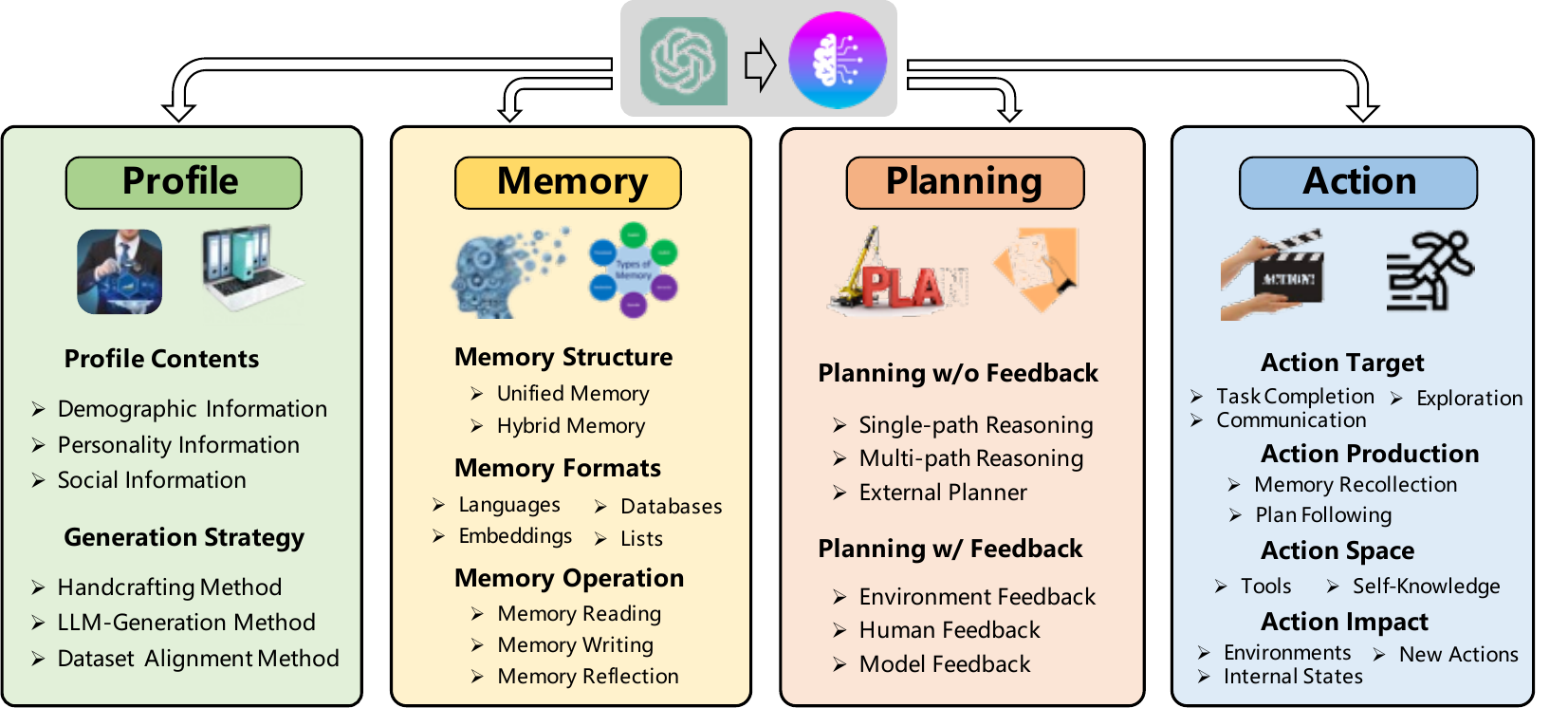}
    }
    \caption{A unified framework for the architecture design of LLM-based autonomous agent.}
    \vspace{-0.cm}
    \label{fs}
\end{figure*}
\section{LLM-based Autonomous Agent Construction}\label{sec:construct}
LLM-based autonomous agents are expected to effectively perform diverse tasks by leveraging the human-like capabilities of LLMs.
In order to achieve this goal, there are two significant aspects: (1) which architecture should be designed to better use LLMs and (2) given the designed architecture, how to enable the agent to acquire capabilities for accomplishing specific tasks.
Within the context of architecture design, we contribute a systematic synthesis of existing research, culminating in a comprehensive unified framework.
% \footnote{Our framework is also inspired by a pioneer work at https://lilianweng.github.io/posts/2023-06-23-agent/}.
As for the second aspect, we summarize the strategies for agent capability acquisition based on whether they fine-tune the LLMs.
Comparing LLM-based autonomous agents to traditional machine learning, architecture design is analogous to defining the network structure, while capability acquisition resembles the process of learning network parameters. 
In the following sections, we explore these two aspects in greater detail.

\subsection{Agent Architecture Design}
Recent advancements in LLMs have demonstrated their great potential to accomplish a wide range of tasks in the form of question-answering (QA).
However, building autonomous agents is far from QA, since they need to fulfill specific roles and autonomously perceive and learn from the environment to evolve themselves like humans.
To bridge the gap between traditional LLMs and autonomous agents, a crucial aspect is to design rational agent architectures to assist LLMs in maximizing their capabilities.
Along this direction, previous work has developed a number of modules to enhance LLMs. 
In this section, we propose a unified framework to summarize these modules. 
Specifically, the overall structure of our framework is illustrated in Figure~\ref{fs}, which is composed of a profiling module, a memory module, a planning module, and an action module.
The purpose of the profiling module is to identify the role of the agent.
The memory and planning modules place the agent into a dynamic environment, enabling it to recall past behaviors and plan future actions.
The action module is responsible for translating the agent's decisions into specific outputs.
Within these modules, the profiling module impacts the memory and planning modules, and collectively, these three modules influence the action module.
In the following, we detail these modules.

\subsubsection{Profiling Module}
Autonomous agents typically perform tasks by assuming specific roles, such as coders, teachers, and domain experts~\cite{qian2023communicative,AgentVerse}.
The profiling module aims to indicate the profiles of the agent roles, which are usually written into the prompt to influence the behavior of the LLM. 
Agent profiles typically encompass basic information such as age, gender, and career~\cite{park2023generative}, as well as psychology information, reflecting the personalities of the agents, and social information, detailing the relationships between agents~\cite{wang2023recagent}.
The choice of information to profile the agent is largely determined by the specific application scenarios.
For instance, if the application aims to study human cognitive process, then the psychology information becomes pivotal.
After identifying the types of profile information, the next important problem is to create specific profiles for the agents. 
Existing literature commonly employs the following three strategies.

\textbf{Handcrafting Method}:
in this method, agent profiles are manually specified. For instance, if one would like to design agents with different personalities, he can use "you are an outgoing person" or "you are an introverted person" to profile the agent. 
The handcrafting method has been leveraged in a lot of previous work to specify the agent profiles.
For example, Generative Agent~\cite{zhang2023building} describes the agent by the information such as name, objectives, and relationships with other agents. 
MetaGPT~\cite{hong2023metagpt}, ChatDev~\cite{qian2023communicative}, and Self-collaboration~\cite{dong2023self} predefine various roles and their corresponding responsibilities in software development, manually assigning distinct profiles to each agent to facilitate collaboration.
PTLLM~\cite{safdari2023personality} aims to explore and quantify personality traits displayed in texts generated by LLMs. 
This method guides LLMs in generating diverse responses by manually defining various agent characters through the use of personality assessment tools such as IPIP-NEO~\cite{johnson2014measuring} and BFI~\cite{john1991big}.
~\cite{deshpande2023toxicity} studies the toxicity of the LLM output by manually prompting LLMs with different roles, such as politicians, journalists and businesspersons.
In general, the handcrafting method is very flexible, since one can assign any profile information to the agents.
However, it can be also labor-intensive, particularly when dealing with a large number of agents.

\textbf{LLM-generation Method}:
in this method, agent profiles are automatically generated based on LLMs.
Typically, it begins by indicating the profile generation rules, elucidating the composition and attributes of the agent profiles within the target population. 
Then, one can optionally specify several seed agent profiles to serve as few-shot examples. 
Finally, LLMs are leveraged to generate all the agent profiles.
For example, RecAgent~\cite{wang2023recagent} first creates seed profiles for a few agents by manually crafting their attributes such as age, gender, personal traits, and movie preferences. Then, it leverages ChatGPT to generate more agent profiles based on the seed information.
This approach significantly reduces the time and effort required to construct agent profiles, particularly for large-scale populations. However, it may lack precise control over the generated profiles, which can result in inconsistencies or deviations from the intended characteristics.

\textbf{Dataset Alignment Method}:
in this method, the agent profiles are obtained from real-world datasets. 
Typically, one can first organize the information about real humans in the datasets into natural language prompts, and then leverage it to profile the agents. 
For instance, in~\cite{argyle2023out}, the authors assign roles to GPT-3 based on the demographic backgrounds (such as race/ethnicity, gender, age, and state of residence) of participants in the American National Election Studies (ANES). They subsequently investigate whether GPT-3 can produce similar results to those of real humans. The dataset alignment method accurately captures the attributes of the real population, thereby making the agent behaviors more meaningful and reflective of real-world scenarios.

\begin{remark}
While most of the previous work leverage the above profile generation strategies independently, we argue that combining them may yield additional benefits.
For example, in order to predict social developments via agent simulation, one can leverage real-world datasets to profile a subset of the agents, thereby accurately reflecting the current social status.
Subsequently, roles that do not exist in the real world but may emerge in the future can be manually assigned to the other agents, enabling the prediction of future social development.
Beyond this example, one can also flexibly combine the other strategies.
The profile module serves as the foundation for agent design, exerting significant influence on the agent memorization, planning, and action procedures.
\end{remark}

% \begin{figure*}[t]
%     \centering
%     \setlength{\fboxrule}{0.pt}
%     \setlength{\fboxsep}{0.pt}
%     \fbox{
%         \includegraphics[width=1.\linewidth]{./FCS-240231-fig2.pdf}
%     }
%     \caption{A unified framework for the architecture design of LLM-based autonomous agent.}
%     \vspace{-0.cm}
%     \label{fs}
% \end{figure*}

\subsubsection{Memory Module}
The memory module plays a very important role in the agent architecture design. 
It stores information perceived from the environment and leverages the recorded memories to facilitate future actions.
The memory module can help the agent to accumulate experiences, self-evolve, and behave in a more consistent, reasonable, and effective manner.
This section provides a comprehensive overview of the memory module, focusing on its structures, formats, and operations.

\textbf{Memory Structures}: LLM-based autonomous agents often draw inspiration from cognitive science research on human memory processes. Human memory follows a general progression from sensory memory that registers perceptual inputs, to short-term memory that maintains information transiently, to long-term memory that consolidates information over extended periods. 
When designing the agent memory structures, researchers take inspiration from these aspects of human memory.
In particular, short-term memory is analogous to the input information within the context window constrained by the transformer architecture. 
Long-term memory resembles the external vector storage that agents can rapidly query and retrieve from as needed. 
In the following, we introduce two commonly used memory structures based on the short-term and long-term memories.

\quad $\bullet$ \textit{Unified Memory}. 
This structure only simulates the human short-term memory, which is usually realized by in-context learning, and the memory information is directly written into the prompts.
% {Atlas}~\cite{izacard2022atlas} stores document memories based on universal dense vectors, which are generated from a dual-encoder model. 
% Augmented LLM \cite{schuurmans2023memory} employs a unified external storage for its memory, which can be accessed via prompts.
% {Voyager}~\cite{wang2023voyager} also utilizes a unified memory architecture, where skills of different complexities are gathered in a central library.
% During code generation, skills can be indexed based on their relevance for matching and retrieval.
% {ChatLog}~\cite{tu2023chatlog} maintains a unified memory stream, which allows the model to retain important historical information and adaptively adjust the agents themselves for different environments.
For example, RLP~\cite{fischer2023reflective} is a conversation agent, which maintains internal states for the speaker and listener.
During each round of conversation, these states serve as LLM prompts, functioning as the agent's short-term memory.
SayPlan~\cite{rana2023sayplan} is an embodied agent specifically designed for task planning. In this agent, the scene graphs and environment feedback serve as the agent's short-term memory, guiding its actions.
CALYPSO~\cite{zhu2023calypso} is an agent designed for the game Dungeons \& Dragons, which can assist Dungeon Masters in the creation and narration of stories. 
Its short-term memory is built upon scene descriptions, monster information, and previous summaries.
DEPS~\cite{wang2023describe} is also a game agent, developed for Minecraft.
The agent initially generates task plans and then utilizes them to prompt LLMs, which in turn produce actions to complete the task. These plans can be deemed as the agent's short-term memory.
In practice, implementing short-term memory is straightforward and can enhance an agent's ability to perceive recent or contextually sensitive behaviors and observations. 
However, the limited context window of LLMs restricts incorporating comprehensive memories into prompts, which can impair agent performance. This challenge necessitates LLMs with larger context windows and the ability to handle extended contexts. Consequently, numerous researchers turn to hybrid memory systems to mitigate this issue.

\quad $\bullet$ \textit{Hybrid Memory}. 
This structure explicitly models the human short-term and long-term memories. 
The short-term memory temporarily buffers recent perceptions, while long-term memory consolidates important information over time.
For instance, Generative Agent~\cite{park2023generative} employs a hybrid memory structure to facilitate agent behaviors.
The short-term memory contains the context information about the agent current situations, while the long-term memory stores the agent past behaviors and thoughts, which can be retrieved according to the current events.
{AgentSims}~\cite{lin2023agentsims} also implements a hybrid memory architecture. 
The information provided in the prompt can be considered as short-term memory.
In order to enhance the storage capacity of memory, the authors propose a long-term memory system that utilizes a vector database, facilitating efficient storage and retrieval.
Specifically, the agent's daily memories are encoded as embeddings and stored in the vector database.
If the agent needs to recall its previous memories, the long-term memory system retrieves relevant information using embedding similarities.
This process can improve the consistency of the agent's behavior.
In {GITM}~\cite{zhu2023ghost}, the short-term memory stores the current trajectory, and the long-term memory saves reference plans summarized from successful prior trajectories. 
Long-term memory provides stable knowledge, while short-term memory allows flexible planning. 
{Reflexion}~\cite{shinn2023reflexion} utilizes a short-term sliding window to capture recent feedback and incorporates persistent long-term storage to retain condensed insights. This combination allows for the utilization of both detailed immediate experiences and high-level abstractions.
SCM~\cite{liang2023unleashing} selectively activates the most relevant long-term knowledge to combine with short-term memory, enabling reasoning over complex contextual dialogues. 
SimplyRetrieve~\cite{ng2023simplyretrieve} utilizes user queries as short-term memory and stores long-term memory using private knowledge bases.
This design enhances the model accuracy while guaranteeing user privacy. 
MemorySandbox~\cite{huang2023memory} implements long-term and short-term memory to store different objects, which can then be accessed throughout various conversations.
Users can create multiple conversations with different agents on the same canvas, facilitating the sharing of memory objects through a simple drag-and-drop interface.
In practice, integrating both short-term and long-term memories can enhance an agent's ability for long-range reasoning and accumulation of valuable experiences, which are crucial for accomplishing tasks in complex environments. 

\begin{remark}
Careful readers may find that there may also exist another type of memory structure, that is, only based on the long-term memory.
However, we find that such type of memory is rarely documented in the literature.
Our speculation is that the agents are always situated in continuous and dynamic environments, with consecutive actions displaying a high correlation.
Therefore, the capture of short-term memory is very important and usually cannot be disregarded.
\end{remark}

\textbf{Memory Formats}: 
In addition to the memory structure, another perspective to analyze the memory module is based on the formats of the memory storage medium, for example, natural language memory or embedding memory. 
Different memory formats possess distinct strengths and are suitable for various applications. 
In the following, we introduce several representative memory formats.

\quad $\bullet$ \textit{Natural Languages}. 
In this format, memory information such as the agent behaviors and observations are directly described using raw natural language.
This format possesses several strengths.
Firstly, the memory information can be expressed in a flexible and understandable manner.
Moreover, it retains rich semantic information that can provide comprehensive signals to guide agent behaviors.
In the previous work, {Reflexion}~\cite{shinn2023reflexion} stores experiential feedback in natural language within a sliding window. 
{Voyager}~\cite{wang2023voyager} employs natural language descriptions to represent skills within the Minecraft game, which are directly stored in memory.

\quad $\bullet$ \textit{Embeddings}. 
In this format, memory information is encoded into embedding vectors, which enhances both retrieval and reading efficiency.
For instance, {MemoryBank}~\cite{zhong2023memorybank} encodes each memory segment as an embedding vector and employs a dual-tower dense retrieval model to efficiently retrieve relevant information from past conversations.

\quad $\bullet$ \textit{Databases}. 
In this format, memory information is stored in databases, allowing the agent to manipulate memories efficiently and comprehensively. 
For example, {ChatDB}~\cite{hu2023chatdb} uses a database as a symbolic memory module. The agent can utilize SQL statements to precisely add, delete, and modify the memory information.

\quad $\bullet$ \textit{Structured Lists}. 
In this format, memory information is organized into lists, and the semantic of memory can be conveyed in an efficient and concise manner.
For instance, {GITM}~\cite{zhu2023ghost} stores action lists for sub-goals in a hierarchical tree structure.
The hierarchical structure explicitly captures the relationships between goals and corresponding plans.
RET-LLM~\cite{modarressi2023ret} initially converts natural language sentences into triplet phrases, and subsequently stores them in memory.

\begin{remark}
Here we only show several representative memory formats, but it is important to note that there are many uncovered ones, such as the programming code used by~\cite{wang2023voyager}.
Moreover, it should be emphasized that these formats are not mutually exclusive; many models incorporate multiple formats to concurrently harness their respective benefits.
A notable example is the memory module of GITM~\cite{zhu2023ghost}, which utilizes a key-value list structure. In this structure, the keys are represented by embedding vectors, while the values consist of raw natural languages.
The use of embedding vectors allows for efficient retrieval of memory records.
By utilizing natural languages, the memory contents become highly comprehensive, enabling more informed agent actions.
\end{remark}
% \textcolor{red}{In SYMBOLICTOM~\cite{sclar2023minding}, each agent maintains a local graph representing the current state of the agent's brief or memory. In the local graph, nodes represent entities and edges represent relationships. Meanwhile, a global graph is also maintained to represent the real world. }
Above, we mainly discuss the internal designs of the memory module. In the following, we turn our focus to memory operations, which are used to interact with external environments.

\textbf{Memory Operations}: 
The memory module plays a critical role in allowing the agent to acquire, accumulate, and utilize significant knowledge by interacting with the environment. 
The interaction between the agent and the environment is accomplished through three crucial memory operations: memory reading, memory writing, and memory reflection.
In the following, we introduce these operations more in detail.

\quad $\bullet$ \textit{Memory Reading}. 
The objective of memory reading is to extract meaningful information from memory to enhance the agent's actions.
For example, using the previously successful actions to achieve similar goals~\cite{zhu2023ghost}.
The key of memory reading lies in how to extract valuable information from history actions.
Usually, there are three commonly used criteria for information extraction, that is, the recency, relevance, and importance~\cite{park2023generative}.
Memories that are more recent, relevant, and important are more likely to be extracted.
Formally, we conclude the following equation from existing literature for memory information extraction:
\begin{equation}
\begin{aligned}
m^* = \arg\max_{m\in M} \left(\alpha s^{rec}(q,m) + \beta s^{rel}(q,m) + \gamma s^{imp}(m)\right),
\end{aligned}
\end{equation}
where $q$ is the query, for example, the task that the agent should address or the context in which the agent is situated.
$M$ is the set of all memories.
$s^{rec}(\cdot)$, $s^{rel}(\cdot)$ and $s^{imp}(\cdot)$ are the scoring functions for measuring the recency, relevance, and importance of the memory $m$, with higher scores indicating more recent, more relevant, and more important memories respectively.
These scoring functions can be implemented using various methods, for example, $s^{rel}(q,m)$ can be calculated using vector similarity measures between query and memory embeddings.
% \footnote{https://lilianweng.github.io/posts/2023-06-23-agent/}. 
It should be noted that $s^{imp}$ only reflects the characters of the memory itself, thus it is unrelated to the query $q$.
$\alpha$, $\beta$ and $\gamma$ are balancing parameters.
By assigning them with different values, one can obtain various memory reading strategies.
For example, by setting $\alpha=\gamma=0$, many studies~\cite{modarressi2023ret,zhu2023ghost,wang2023voyager,fischer2023reflective} only consider the relevance score $s^{rel}$ for memory reading.
By assigning $\alpha=\beta=\gamma=1.0$, \cite{park2023generative} equally weights all the above three metrics to extract information from memory.

\quad $\bullet$ \textit{Memory Writing}. 
The purpose of memory writing is to store information about the perceived environment in memory.
Storing valuable information in memory provides a foundation for retrieving informative memories in the future, enabling the agent to act more efficiently and rationally.
During the memory writing process, there are two potential problems that should be carefully addressed.
On one hand, it is crucial to address how to store information that is similar to existing memories (\emph{i.e.}, memory duplicated).
On the other hand, it is important to consider how to remove information when the memory reaches its storage limit (\emph{i.e.}, memory overflow).
In the following, we discuss these problems more in detail.
(1) \textit{Memory Duplicated}. 
To incorporate similar information, people have developed various methods for integrating new and previous records.
For instance, in~\cite{zhu2023ghost}, the successful action sequences related to the same sub-goal are stored in a list.
Once the size of the list reaches N(=5), all the sequences in it are condensed into a unified plan solution using LLMs.
The original sequences in the memory are replaced with the newly generated one.
{Augmented LLM} \cite{schuurmans2023memory} aggregates duplicate information via count accumulation, avoiding redundant storage.
% {Reflexion} \cite{shinn2023reflexion} consolidates related feedback into high-level insights, and replacing raw experiences.
(2) \textit{Memory Overflow}. 
In order to write information into the memory when it is full, people design different methods to delete existing information to continue the memorizing process.
% Additional strategies can be designed to discard memories, such as considering their importance or dropping them randomly.
For example, in {ChatDB}~\cite{hu2023chatdb}, memories can be explicitly deleted based on user commands.
{RET-LLM}~\cite{modarressi2023ret} uses a fixed-size buffer for memory, overwriting the oldest entries in a first-in-first-out (FIFO) manner.

\quad $\bullet$ \textit{Memory Reflection}. 
Memory reflection emulates humans' ability to witness and evaluate their own cognitive, emotional, and behavioral processes.
When adapted to agents, the objective is to provide agents with the capability to independently summarize and infer more abstract, complex and high-level information.
More specifically, in Generative Agent~\cite{park2023generative}, the agent has the capability to summarize its past experiences stored in memory into broader and more abstract insights.
To begin with, the agent generates three key questions based on its recent memories.
Then, these questions are used to query the memory to obtain relevant information.
Building upon the acquired information, the agent generates five insights, which reflect the agent high-level ideas.
For example, the low-level memories ``Klaus Mueller is writing a research paper'', ``Klaus Mueller is engaging with a librarian to further his research'', and ``Klaus Mueller is conversing with Ayesha Khan
about his research'' can induce the high-level insight ``Klaus Mueller is dedicated to his research''.
In addition, the reflection process can occur hierarchically, meaning that the insights can be generated based on existing insights.
In GITM~\cite{zhu2023ghost}, the actions that successfully accomplish the sub-goals are stored in a list.
When the list contains more than five elements, the agent summarizes them into a common and abstract pattern and replaces all the elements.
In ExpeL~\cite{zhao2023expel}, two approaches are introduced for the agent to acquire reflection. Firstly, the agent compares successful or failed trajectories within the same task. Secondly, the agent learns from a collection of successful trajectories to gain experiences.

A significant distinction between traditional LLMs and the agents is that the latter must possess the capability to learn and complete tasks in dynamic environments.
If we consider the memory module as responsible for managing the agents' past behaviors, it becomes essential to have another significant module that can assist the agents in planning their future actions.
In the following, we present an overview of how researchers design the planning module.

\begin{figure*}[t]
    \centering
    \setlength{\fboxrule}{0.pt}
    \setlength{\fboxsep}{0.pt}
    \fbox{
        \includegraphics[width=.9\linewidth]{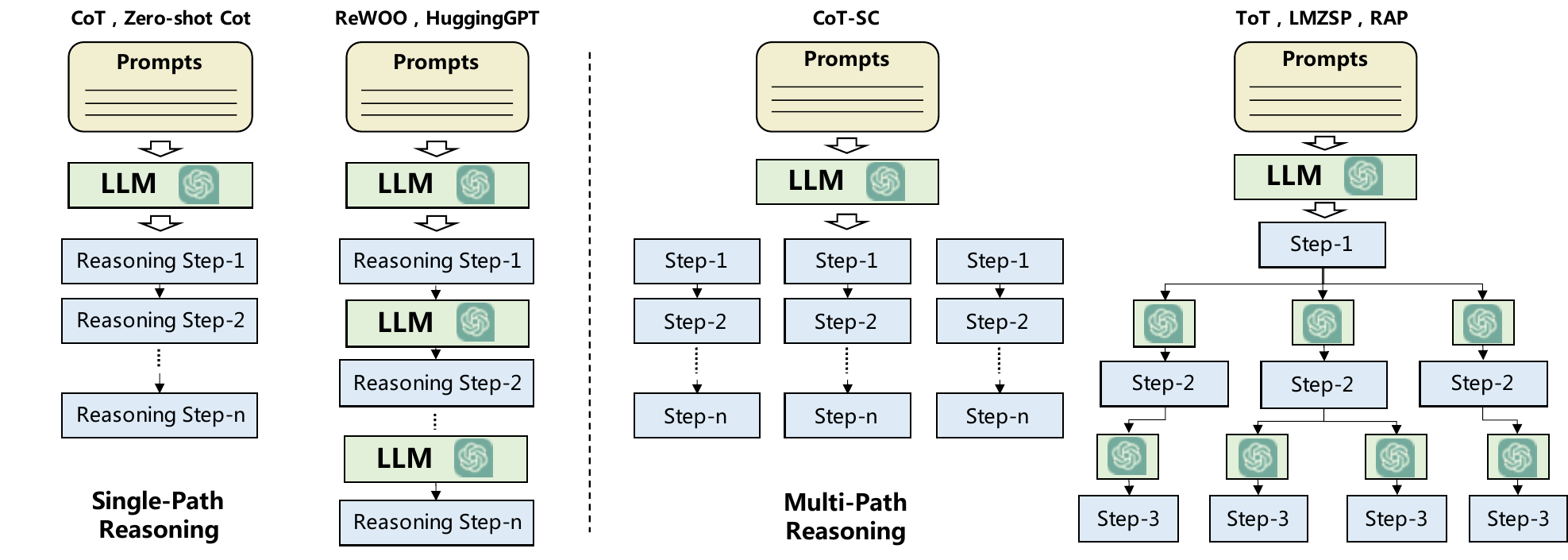}
    }
    \caption{Comparison between the strategies of single-path and multi-path reasoning. LMZSP is the model proposed in~\cite{huang2022language}.}
    \label{pl}
    \vspace{-0.cm}
\end{figure*}

\subsubsection{Planning Module}
When faced with a complex task, humans tend to deconstruct it into simpler subtasks and solve them individually.
The planning module aims to empower the agents with such human capability, which is expected to make the agent behave more reasonably, powerfully, and reliably.
In specific, we summarize existing studies based on whether the agent can receive feedback in the planing process, which are detailed as follows:

\textbf{Planning without Feedback}: 
In this method, the agents do not receive feedback that can influence its future behaviors after taking actions.
In the following, we present several representative strategies.

\quad $\bullet$ \textit{Single-path Reasoning}. 
In this strategy, the final task is decomposed into several intermediate steps. 
These steps are connected in a cascading manner, with each step leading to only one subsequent step.
LLMs follow these steps to achieve the final goal.
Specifically, Chain of Thought (CoT)~\cite{wei2022chain} proposes inputting reasoning steps for solving complex problems into the prompt. These steps serve as examples to inspire LLMs to plan and act in a step-by-step manner.
In this method, the plans are created based on the inspiration from the examples in the prompts.
Zero-shot-CoT~\cite{kojima2022large} enables LLMs to generate task reasoning processes by prompting them with trigger sentences like "think step by step".
Unlike CoT, this method does not incorporate reasoning steps as examples in the prompts. 
Re-Prompting~\cite{raman2022planning} involves checking whether each step meets the necessary prerequisites before generating a plan. 
If a step fails to meet the prerequisites, it introduces a prerequisite error message and prompts the LLM to regenerate the plan.
ReWOO~\cite{xu2023rewoo} introduces a paradigm of separating plans from external observations, where the agents first generate plans and obtain observations independently, and then combine them together to derive the final results.
HuggingGPT~\cite{shen2023hugginggpt} first decomposes the task into many sub-goals, and then solves each of them based on Huggingface.
Different from CoT and Zero-shot-CoT, which outcome all the reasoning steps in a one-shot manner, ReWOO and HuggingGPT produce the results by accessing LLMs multiply times.
% SWIFTSAGE~\cite{lin2023swiftsage} uses a combination of fast and slow thinking for planning in complex interactive tasks, inspired by human cognitive dual-process theory~\cite{evans2013dual}.
SWIFTSAGE~\cite{lin2023swiftsage}, inspired by the dual-process theory of human cognition~\cite{evans2013dual}, combines the power of both SWIFT and SAGE modules for planning in complex interactive tasks. The SWIFT module provides quick responses based on learned patterns, while the SAGE module, using large language models, conducts in-depth planning by asking key questions and generating action sequences to ensure successful task completion.

\quad $\bullet$ \textit{Multi-path Reasoning}. 
In this strategy, the reasoning steps for generating the final plans are organized into a tree-like structure.
Each intermediate step may have multiple subsequent steps.
This approach is analogous to human thinking, as individuals may have multiple choices at each reasoning step.
In specific, Self-consistent CoT (CoT-SC)~\cite{wang2022self} believes that each complex problem has multiple ways of thinking to deduce the final answer. 
Thus, it starts by employing CoT to generate various reasoning paths and corresponding answers. Subsequently, the answer with the highest frequency is chosen as the final output.
Tree of Thoughts (ToT)~\cite{yao2023tree} is designed to generate plans using a tree-like reasoning structure. In this approach, each node in the tree represents a "thought," which corresponds to an intermediate reasoning step. The selection of these intermediate steps is based on the evaluation of LLMs. The final plan is generated using either the breadth-first search (BFS) or depth-first search (DFS) strategy.
Comparing with CoT-SC, which generates all the planed steps together, ToT needs to query LLMs for each reasoning step.
In RecMind~\cite{wang2023recmind}, the authors designed a self-inspiring mechanism, where the discarded historical information in the planning process is also leveraged to derive new reasoning steps.
In GoT~\cite{besta2023graph}, the authors expand the tree-like reasoning structure in ToT to graph structures, resulting in more powerful prompting strategies.
In AoT~\cite{sel2023algorithm}, the authors design a novel method to enhance the reasoning processes of LLMs by incorporating algorithmic examples into the prompts.
Remarkably, this method only needs to query LLMs for only one or a few times.
In~\cite{huang2022language}, the LLMs are leveraged as zero-shot planners.
At each planning step, they first generate multiple possible next steps, and then determine the final one based on their distances to admissible actions.
~\cite{gramopadhye2023generating} further improves ~\cite{huang2022language} by incorporating examples that are similar to the queries in the prompts.
RAP~\cite{hao2023reasoning} builds a world model to simulate the potential benefits of different plans based on Monte Carlo Tree Search (MCTS), and then, the final plan is generated by aggregating multiple MCTS iterations.
To enhance comprehension, we provide an illustration comparing the strategies of single-path and multi-path reasoning in Figure~\ref{pl}.

% \begin{figure*}[t]
%     \centering
%     \setlength{\fboxrule}{0.pt}
%     \setlength{\fboxsep}{0.pt}
%     \fbox{
%         \includegraphics[width=1.\linewidth]{./FCS-240231-fig3.pdf}
%     }
%     \caption{Comparison between the strategies of single-path and multi-path reasoning. LMZSP is the model proposed in~\cite{huang2022language}.}
%     \label{pl}
%     \vspace{-0.cm}
% \end{figure*}

\quad $\bullet$ \textit{External Planner}. 
Despite the demonstrated power of LLMs in zero-shot planning, effectively generating plans for domain-specific problems remains highly challenging.
To address this challenge, researchers turn to external planners.
These tools are well-developed and employ efficient search algorithms to rapidly identify correct, or even optimal, plans.
In specific, LLM+P~\cite{liu2023llmp+} first transforms the task descriptions into formal Planning Domain Definition Languages (PDDL), and then it uses an external planner to deal with the PDDL. Finally, the generated results are transformed back into natural language by LLMs.
Similarly, LLM-DP~\cite{dagan2023dynamic} utilizes LLMs to convert the observations, the current world state, and the target objectives into PDDL. Subsequently, this transformed data is passed to an external planner, which efficiently determines the final action sequence.
CO-LLM~\cite{zhang2023building} demonstrates that LLMs is good at generating high-level plans, but struggle with low-level control. 
To address this limitation, a heuristically designed external low-level planner is employed to effectively execute actions based on high-level plans.

\textbf{Planning with Feedback}: 
In many real-world scenarios, the agents need to make long-horizon planning to solve complex tasks.
When facing these tasks, the above planning modules without feedback can be less effective due to the following reasons:
firstly, generating a flawless plan directly from the beginning is extremely difficult as it needs to consider various complex preconditions. 
As a result, simply following the initial plan often leads to failure. 
Moreover, the execution of the plan may be hindered by unpredictable transition dynamics, rendering the initial plan non-executable. 
Simultaneously, when examining how humans tackle complex tasks, we find that individuals may iteratively make and revise their plans based on external feedback.
To simulate such human capability, researchers have designed many planning modules, where the agent can receive feedback after taking actions.
The feedback can be obtained from environments, humans, and models, which are detailed in the following.

\quad $\bullet$ \textit{Environmental Feedback}. 
This feedback is obtained from the objective world or virtual environment. 
For instance, it could be the game's task completion signals or the observations made after the agent takes an action.
In specific, 
ReAct~\cite{yao2022react} proposes constructing prompts using thought-act-observation triplets.
The thought component aims to facilitate high-level reasoning and planning for guiding agent behaviors.
The act represents a specific action taken by the agent.
The observation corresponds to the outcome of the action, acquired through external feedback, such as search engine results.
The next thought is influenced by the previous observations, which makes the generated plans more adaptive to the environment.
Voyager~\cite{wang2023voyager} makes plans by incorporating three types of environment feedback including the intermediate progress of program execution, the execution error and self-verification results.
These signals can help the agent to make better plans for the next action.
Similar to Voyager, Ghost~\cite{zhu2023ghost} also incorporates feedback into the reasoning and action taking processes.
This feedback encompasses the environment states as well as the success and failure information for each executed action.
{SayPlan~\cite{rana2023sayplan} leverages environmental feedback derived from a scene graph simulator to validate and refine its strategic formulations. 
This simulator is adept at discerning the outcomes and state transitions subsequent to agent actions, facilitating SayPlan's iterative recalibration of its strategies until a viable plan is ascertained.}
In DEPS~\cite{wang2023describe},  the authors argue that solely providing information about the completion of a task is often inadequate for correcting planning errors.
Therefore, they propose informing the agent about the detail reasons for task failure, allowing them to more effectively revise their plans.
LLM-Planner~\cite{song2023llmplanner} introduces a grounded re-planning algorithm that dynamically updates plans generated by LLMs when encountering object mismatches and unattainable plans during task completion. 
Inner Monologue~\cite{huang2022inner} provides three types of feedback to the agent after it takes actions:
(1) whether the task is successfully completed, (2) passive scene descriptions, and (3) active scene descriptions.
The former two are generated from the environments, which makes the agent actions more reasonable.

% REX~\cite{murthy2023rex} employs the UCB score or simple rewards to influence the action generation and selection of AI agents, enabling a balance between exploration and exploitation.

% Introspective Tips~\cite{chen2023introspective} allows LLM to introspect through the history of the environmental feedback.
% LLM-Planner~\cite{song2023llmplanner} introduces a grounded re-planning algorithm that dynamically updates plans generated by LLMs when encountering object mismatches and unattainable plans during task completion.
% In Progprompt~\cite{singh2022progprompt}, assertions are incorporated into the generated script to provide environment state feedback, allowing for error recovery in case the action's preconditions are not satisfied.
% In summary, environmental feedback serves as a direct indicator of planning success or failure, thereby enhancing the efficiency of closed-loop planning.

\quad $\bullet$ \textit{Human Feedback}. 
In addition to obtaining feedback from the environment, directly interacting with humans is also a very intuitive strategy to enhance the agent planning capability.
The human feedback is a subjective signal.
It can effectively make the agent align with the human values and preferences, and also help to alleviate the hallucination problem.
% In SayCan~\cite{ahn2022can}, the agent employs a two-level structure for planning. 
% It leverages human feedback to train a reinforcement learning model that enables accurate action execution at the low-level skills.
In Inner Monologue~\cite{huang2022inner}, the agent aims to perform high-level natural language instructions in a 3D visual environment.
It is given the capability to actively solicit feedback from humans regarding scene descriptions.
Then, the agent incorporates the human feedback into its prompts, enabling more informed planning and reasoning.
In the above cases, we can see, different types of feedback can be combined to enhance the agent planning capability.
For example, Inner Monologue~\cite{huang2022inner} collects both environment and human feedback to facilitate the agent plans.

\quad $\bullet$ \textit{Model Feedback}. 
Apart from the aforementioned environmental and human feedback, which are external signals, researchers have also investigated the utilization of internal feedback from the agents themselves.
This type of feedback is usually generated based on pre-trained models.
In specific,~\cite{madaan2023self} proposes a self-refine mechanism.
This mechanism consists of three crucial components: output, feedback, and refinement. Firstly, the agent generates an output. Then, it utilizes LLMs to provide feedback on the output and offer guidance on how to refine it. At last, the output is improved by the feedback and refinement.
This output-feedback-refinement process iterates until reaching some desired conditions.
SelfCheck~\cite{miao2023selfcheck} allows agents to examine and evaluate their reasoning steps generated at various stages. They can then correct any errors by comparing the outcomes.
InterAct~\cite{chen2023interact} uses different language models (such as ChatGPT and InstructGPT) as auxiliary roles, such as checkers and sorters, to help the main language model avoid erroneous and inefficient actions.
ChatCoT~\cite{chen2023chatcot} utilizes model feedback to improve the quality of its reasoning process. 
The model feedback is generated by an evaluation module that monitors the agent reasoning steps.
Reflexion~\cite{shinn2023reflexion} is developed to enhance the agent's planning capability through detailed verbal feedback.
In this model, the agent first produces an action based on its memory, and then, the evaluator generates feedback by taking the agent trajectory as input.
In contrast to previous studies, where the feedback is given as a scalar value, this model leverages LLMs to provide more detailed verbal feedback, which can provide more comprehensive supports for the agent plans.

\begin{remark}
In conclusion, the implementation of planning module without feedback is relatively straightforward. However, it is primarily suitable for simple tasks that only require a small number of reasoning steps.
Conversely, the strategy of planning with feedback needs more careful designs to handle the feedback.
Nevertheless, it is considerably more powerful and capable of effectively addressing complex tasks that involve long-range reasoning.
\end{remark}

\vspace{-0.2cm}
\subsubsection{Action Module}
The action module is responsible for translating the agent's decisions into specific outcomes. This module is located at the most downstream position and directly interacts with the environment. It is influenced by the profile, memory, and planning modules. This section introduces the action module from four perspectives: (1) Action goal: what are the intended outcomes of the actions?  (2) Action production: how are the actions generated? (3) Action space: what are the available actions? (4) Action impact: what are the consequences of the actions? 
Among these perspectives, the first two focus on the aspects preceding the action ("before-action" aspects), the third focuses on the action itself ("in-action" aspect), and the fourth emphasizes the impact of the actions ("after-action" aspect).

\textbf{Action Goal}:
The agent can perform actions with various objectives.
Here, we present several representative examples:
(1) \textit{Task Completion}.
In this scenario, the agent's actions are aimed at accomplishing specific tasks, such as crafting an iron pickaxe in Minecraft~\cite{wang2023voyager} or completing a function in software development~\cite{qian2023communicative}.
These actions usually have well-defined objectives, and each action contributes to the completion of the final task.
Actions aimed at this type of goal are very common in existing literature.
(2) \textit{Communication}.
In this case, the actions are taken to communicate with the other agents or real humans for sharing information or collaboration.
For example, the agents in ChatDev~\cite{qian2023communicative} may communicate with each other to collectively accomplish software development tasks.
In Inner Monologue~\cite{huang2022inner}, the agent actively engages in communication with humans and adjusts its action strategies based on human feedback.
(3) \textit{Environment Exploration}. 
In this example, the agent aims to explore unfamiliar environments to expand its perception and strike a balance between exploring and exploiting.
For instance, the agent in Voyager~\cite{wang2023voyager} may explore unknown skills in their task completion process and continually refine the skill execution code based on environment feedback through trial and error.

\textbf{Action Production}:
Different from ordinary LLMs, where the model input and output are directly associated, the agent may take actions via different strategies and sources.
In the following, we introduce two types of commonly used action production strategies.
(1) \textit{Action via Memory Recollection}. 
In this strategy, the action is generated by extracting information from the agent memory according to the current task.
The task and the extracted memories are used as prompts to trigger the agent actions.
For example, in Generative Agents~\cite{park2023generative}, the agent maintains a memory stream, and before taking each action, it retrieves recent, relevant and important information from the memory steam to guide the agent actions.
In GITM~\cite{zhu2023ghost}, in order to achieve a low-level sub-goal, the agent queries its memory to determine if there are any successful experiences related to the task. If similar tasks have been completed previously, the agent invokes the previously successful actions to handle the current task directly.
In collaborative agents such as ChatDev~\cite{qian2023communicative} and MetaGPT~\cite{hong2023metagpt}, different agents may communicate with each other.
In this process, the conversation history in a dialog is remembered in the agent memories.
Each utterance generated by the agent is influenced by its memory.
(2) \textit{Action via Plan Following}.
In this strategy, the agent takes actions following its pre-generated plans.
For instance, in DEPS~\cite{wang2023describe}, for a given task, the agent first makes action plans. If there are no signals indicating plan failure, the agent will strictly adhere to these plans.
In GITM~\cite{zhu2023ghost}, the agent makes high-level plans by decomposing the task into many sub-goals.
Based on these plans, the agent takes actions to solve each sub-goal sequentially to complete the final task.

\textbf{Action Space}: 
Action space refers to the set of possible actions that can be performed by the agent. 
In general, we can roughly divide these actions into two classes: (1) external tools and (2) internal knowledge of the LLMs.
In the following, we introduce these actions more in detail.

\quad $\bullet$ \textit{External Tools}. 
While LLMs have been demonstrated to be effective in accomplishing a large amount of tasks, they may not work well for the domains which need comprehensive expert knowledge.
In addition, LLMs may also encounter hallucination problems, which are hard to be resolved by themselves.
To alleviate the above problems, the agents are empowered with the capability to call external tools for executing action.
In the following, we present several representative tools which have been exploited in the literature.

(1) \textit{APIs}. 
Leveraging external APIs to complement and expand action space is a popular paradigm in recent years.
For example, 
HuggingGPT~\cite{shen2023hugginggpt} integrates HuggingFace’s vast model ecosystem to tackle complex user tasks.
Similarly, WebGPT~\cite{nakano2021webgpt} proposes to automatically generate queries to extract relevant content from external web pages when responding to user request.
TPTU~\cite{ruan2023tptu} explores the potential of LLMs to address intricate tasks through strategic task planning and API-based tools.
Gorilla~\cite{patil2023gorilla} introduces a fine-tuned LLM capable of generating precise input arguments for API calls, effectively mitigating hallucination issues during external API usage. ToolFormer~\cite{schick2023toolformer} employs self-supervised learning to determine when and how to invoke external tools, using demonstrations of tool APIs for training. API-Bank~\cite{li2023api} offers a comprehensive benchmark with a diverse collection of API tools to systematically evaluate tool-augmented LLMs, alongside robust training datasets designed to enhance their integration capabilities.
ToolLLaMA~\cite{qin2023toolllm} proposes a tool-use framework encompassing data collection, training, and evaluation, with the resulting fine-tuned model excelling across a wide array of APIs.
RestGPT~\cite{song2023restgpt} connects LLMs with RESTful APIs, which follow widely accepted standards for web services development, making the resulting program more compatible with real-world applications.
TaskMatrix.AI~\cite{liang2023taskmatrix} connects LLMs with an extensive ecosystem of APIs to support task execution. 
At its core lies a multimodal conversational foundational model that interacts with users, understands their goals and context, and then produces executable code for particular tasks.
In essence, these intelligent agents strategically harness external APIs as versatile tools, systematically expanding their action space and transcending the inherent limitations of traditional language models by integrating diverse computational capabilities.

(2) \textit{Databases \& Knowledge Bases}. 
Integrating external database or knowledge base enables agents to obtain specific domain information for generating more realistic actions. 
For example, ChatDB~\cite{hu2023chatdb} employs SQL statements to query databases, facilitating actions by the agents in a logical manner. 
MRKL~\cite{karpas2022mrkl} and OpenAGI~\cite{ge2024openagi} incorporate various expert systems such as knowledge bases and planners to access domain-specific information.

(3) \textit{External Models}. 
Previous studies often utilize external models to expand the range of possible actions.
In comparison to APIs, external models typically handle more complex tasks.
Each external model may correspond to multiple APIs.
For example, ViperGPT~\cite{suris2023vipergpt} firstly uses Codex, which is implemented based on language model, to generate Python code from text descriptions, and then executes the code to complete the given tasks. 
ChemCrow~\cite{bran2023chemcrow} is an LLM-based chemical agent designed to perform tasks in organic synthesis, drug discovery, and material design. It utilizes seventeen expert-designed models to assist its operations.
MM-REACT~\cite{yang2023mm} integrates various external models, such as VideoBERT for video summarization, X-decoder for image generation,  and SpeechBERT for audio processing, enhancing its capability in diverse multimodal scenarios.

\quad $\bullet$ \textit{Internal Knowledge}. 
In addition to utilizing external tools, many agents rely solely on the internal knowledge of LLMs to guide their actions.
We now present several crucial capabilities of LLMs that can support the agent to behave reasonably and effectively.
(1) \textit{Planning Capability}. 
Previous work has demonstrated that LLMs can be used as decent planners to decompose complex tasks into simpler ones~\cite{wei2022chain}.
Such a capability of LLMs can be even triggered without incorporating examples in the prompts~\cite{kojima2022large}.
Building on the planning capability of LLMs, DEPS~\cite{wang2023describe} develops a Minecraft agent, which can solve complex tasks via sub-goal decomposition.
Similar agents like GITM~\cite{zhu2023ghost} and Voyager~\cite{wang2023voyager} also heavily rely on the planning capability of LLMs to successfully complete various tasks.
(2) \textit{Conversation Capability}. 
LLMs can usually generate high-quality conversations.
This capability enables agents to behave more like humans.
In the previous work, many agents take actions based on the strong conversation capability of LLMs.
For example, in ChatDev~\cite{qian2023communicative}, different agents can discuss the software development process and reflect on their own behaviors.
In RLP~\cite{fischer2023reflective}, the agent can communicate with the listeners based on their potential feedback on the agent's utterance.
(3) \textit{Common Sense Understanding Capability}. 
Another important capability of LLMs is that they can well comprehend human common sense.
Based on this capability, many agents can simulate human daily life and make human-like decisions.
For example, in Generative Agent~\cite{park2023generative}, the agent can accurately understand its current state, the surrounding environment, and summarize high-level ideas based on basic observations.
Without the common sense understanding capability of LLMs, these behaviors cannot be reliably simulated.
Similar conclusions may also apply to RecAgent~\cite{wang2023recagent} and S3~\cite{gao2023s}, where the agents focus on simulating user social behaviors.

\textbf{Action Impact}:
Action impact refers to the consequences of an agent’s actions. While the range of possible impacts is vast, we highlight a few key examples for clarity:
(1) \textit{Changing Environments.}
Agents can directly alter environment states by actions, such as moving their positions, collecting items, constructing buildings, etc.
For instance, in GITM~\cite{zhu2023ghost} and Voyager~\cite{wang2023voyager}, the environments are changed by the actions of the agents in their task completion process.
Specifically, when an agent collects resources—such as harvesting three pieces of wood—the resources disappear from the environment.
(2) \textit{Altering Internal States.}
Actions taken by the agent can also change the agent itself, including updating memories, forming new plans, acquiring novel knowledge, and more.
For example, in Generative Agents~\cite{park2023generative}, memory streams are updated after performing actions within the system.
Similarly, SayCan~\cite{ahn2022can} enables agents to take actions to update understandings of the environment.
(3) \textit{Triggering New Actions.}
In task completion processes, one action often leads to subsequent actions. For example, in Voyager~\cite{wang2023voyager}, once the agent has gathered the necessary resources, it triggers the construction of buildings.
% DEPS~\cite{wang2023describe} decomposes plans into sequential sub-goals, with each sub-goal potentially triggering the next one.

\setlength{\textfloatsep}{0.8cm}
\begin{table*}[ht!]
    \centering
    \caption{
    % Summary of the construction strategies of representative agents (more agents can be seen on https://github.com/Paitesanshi/LLM-Agent-Survey). 
    For the profile module, we use \whitecircle{1}, \whitecircle{2} and \whitecircle{3} to represent the handcrafting method, LLM-generation method, and dataset alignment method, respectively.
    For the memory module, we focus on the implementation strategies for memory operation and memory structure.
    For memory operation, we use \whitecircle{1} and \whitecircle{2} to indicate that the model only has read/write operations and has read/write/reflection operations, respectively.
    For memory structure, we use \whitecircle{1} and \whitecircle{2} to represent unified and hybrid memories, respectively.
    For the planning module, we use \whitecircle{1} and \whitecircle{2} to represent planning w/o feedback and w/ feedback, respectively. 
    For the action module, we use \whitecircle{1} and \whitecircle{2} to represent that the model does not use tools and use tools, respectively.
    For the agent \underline{c}apability \underline{a}cquisition (CA) strategy, we use \whitecircle{1} and \whitecircle{2} to represent the methods with and without fine-tuning, respectively.
    ``-'' indicates that the corresponding content is not explicitly discussed in the paper. 
    }
    \vspace{0.1cm}
    \renewcommand\arraystretch{0.85}
    \scalebox{.93}{
        \begin{tabular}{p{4cm}p{1.8cm}<{\centering}p{2.4cm}<{\centering}p{1.8cm}<{\centering}p{1.5cm}<{\centering}p{1.2cm}<{\centering}p{1.2cm}<{\centering}p{1.5cm}<{\centering}}
            \hline \hline
            \multirow{2}{*}{Model}&\multirow{2}{*}{Profile}  & \multicolumn{2}{c}{Memory} & \multirow{2}{*}{Planning}  & \multirow{2}{*}{Action} & \multirow{2}{*}{CA}&\multirow{2}{*}{Time}            \\\cline{3-4}

            & & Operation & Structure &&&&  \\
            \hline
            WebGPT~\cite{nakano2021webgpt}&- &- &- &-   &\whitecircle{2} &\whitecircle{1} &12/2021   \\
            SayCan~\cite{ahn2022can}&-&-&-&\whitecircle{1}&\whitecircle{1}&\whitecircle{2}&04/2022\\
            MRKL~\cite{karpas2022mrkl}&- &- &- &\whitecircle{1}  &\whitecircle{2} &- &05/2022   \\
            Inner Monologue~\cite{huang2022inner}&-&-&-&\whitecircle{2}&\whitecircle{1}&\whitecircle{2}&07/2022\\
           Social Simulacra~\cite{park2022social}& \whitecircle{2} & -  & - & - &\whitecircle{1} &- &08/2022  \\
           % Silicon Samples\cite{argyle2023out}& \whitecircle{3} & -  & - & - &\whitecircle{2} &- &09/2022  \\
           ReAct~\cite{yao2022react}& - & -  & - & \whitecircle{2} &\whitecircle{2} &\whitecircle{1} &10/2022             \\
           % REPLUG\cite{shi2023replug} &- &\whitecircle{2} &\whitecircle{1} &-  &\whitecircle{1} &- &01/2023   \\
           MALLM~\cite{schuurmans2023memory} &- &\whitecircle{1} &\whitecircle{2} &-  &\whitecircle{1} &- &01/2023   \\
           DEPS~\cite{wang2023describe}& - & -  & - & \whitecircle{2}&\whitecircle{1} &\whitecircle{2} &02/2023             \\
           Toolformer~\cite{schick2023toolformer}&- &- &- &\whitecircle{1}  &\whitecircle{2} &\whitecircle{1} &02/2023   \\
           Reflexion~\cite{shinn2023reflexion}& - & \whitecircle{2}  & \whitecircle{2} & \whitecircle{2} &\whitecircle{1} &\whitecircle{2} &03/2023             \\
           %Self-Refine\cite{madaan2023self}& - & -  & - & \whitecircle{2} &\whitecircle{1} &? &03/2023             \\
           %DERA\cite{nair2023dera} &\whitecircle{1} &\whitecircle{2} &\whitecircle{2} &\whitecircle{2}  &\whitecircle{1} &\whitecircle{2} &03/2023   \\
           %Colt5\cite{ainslie2023colt5} &- &\whitecircle{2} &\whitecircle{2} &-  &\whitecircle{1} &- &03/2023   \\
           %DERA\cite{nair2023dera}&\whitecircle{2} &- &- &\whitecircle{2}  &\whitecircle{1} &- &03/2023   \\
           CAMEL~\cite{li2023camel} &\whitecircle{1} \whitecircle{2} &- &- &\whitecircle{2}  &\whitecircle{1} &- &03/2023   \\
           API-Bank~\cite{li2023api}& - & -  & - & \whitecircle{2}&\whitecircle{2} &\whitecircle{2} &04/2023             \\
           ViperGPT~\cite{suris2023vipergpt}&- &- &- &-  &\whitecircle{2} &- &03/2023   \\
           HuggingGPT~\cite{shen2023hugginggpt}&- &\whitecircle{1} &\whitecircle{1} &\whitecircle{1}  &\whitecircle{2} &- &03/2023   \\
           Generative Agents~\cite{park2023generative}& \whitecircle{1} & \whitecircle{2}  & \whitecircle{2} & \whitecircle{2} &\whitecircle{1} &- &04/2023             \\
            LLM+P~\cite{liu2023llmp+}& - & -  & - & \whitecircle{1} &\whitecircle{1} &- &04/2023             \\
            ChemCrow~\cite{bran2023chemcrow}&- &- &- &\whitecircle{2}  &\whitecircle{2} &- &04/2023   \\
            %Refiner\cite{paul2023refiner}&- &- &- &\whitecircle{2}  &\whitecircle{2} &\whitecircle{1} &04/2023   \\
           OpenAGI~\cite{ge2024openagi}& - & -  & - & \whitecircle{2} & \whitecircle{2} & \whitecircle{1} &04/2023                  \\
           AutoGPT~\cite{Auto-gpt}& - & \whitecircle{1}  &  \whitecircle{2} & \whitecircle{2} & \whitecircle{2} & \whitecircle{2}&04/2023                  \\
           SCM~\cite{liang2023unleashing} &- &\whitecircle{2} &\whitecircle{2} &-  &\whitecircle{1} &- &04/2023   \\
           %SayCan\cite{ahn2022can} &- &\whitecircle{2} &\whitecircle{2} &\whitecircle{2}  &\whitecircle{1} &\whitecircle{1} &04/2023   \\
           % IGLU\cite{mehta2023improving} &- &- &- &\whitecircle{2}  &\whitecircle{1} &\whitecircle{2} &04/2023   \\
           % ChatLog\cite{tu2023chatlog} &\- &\whitecircle{1} &\whitecircle{1} &\whitecircle{1}  &\whitecircle{1} &\whitecircle{1} &04/2023   \\
           Socially Alignment~\cite{liu2023training}& - & \whitecircle{1}  &  \whitecircle{2} & - & \whitecircle{1} & \whitecircle{1}&05/2023  \\
           GITM~\cite{zhu2023ghost}& - & \whitecircle{2}  & \whitecircle{2} & \whitecircle{2} &\whitecircle{1} &\whitecircle{2} &05/2023             \\
           Voyager~\cite{wang2023voyager}& - & \whitecircle{2}  & \whitecircle{2} & \whitecircle{2} &\whitecircle{1} &\whitecircle{2} &05/2023             \\
           Introspective Tips~\cite{chen2023introspective}& - & -  & - & \whitecircle{2} &\whitecircle{1} &\whitecircle{2} &05/2023    \\     
           % Gorilla\cite{patil2023gorilla}&- &- &- &\whitecircle{2}  &\whitecircle{2} &- &05/2023   \\
           RET-LLM~\cite{modarressi2023ret} &- &\whitecircle{1} &\whitecircle{2} &-  &\whitecircle{1} &\whitecircle{1} &05/2023   \\
           % Self-Notes\cite{lanchantin2023learning} &- &\whitecircle{2} &\whitecircle{2} &- &\whitecircle{1} &- &05/2023   \\
           %SocialAGI\cite{fischer2023reflective} &\whitecircle{2} &\whitecircle{2} &\whitecircle{2} &\whitecircle{2}  &\whitecircle{2} &\whitecircle{2} &05/2023   \\
           ChatDB~\cite{hu2023chatdb} &- &\whitecircle{1} &\whitecircle{2} &\whitecircle{2}  &\whitecircle{2} &- &06/2023   \\
            $S^3$~\cite{gao2023s}& \whitecircle{3} &  \whitecircle{2} & \whitecircle{2} & - &\whitecircle{1} &- &07/2023             \\
           ChatDev~\cite{qian2023communicative} &\whitecircle{1} &\whitecircle{2} &\whitecircle{2} &\whitecircle{2}  &\whitecircle{1} &\whitecircle{2} &07/2023   \\
           ToolLLM~\cite{qin2023toolllm}&- &- &- &\whitecircle{2}  &\whitecircle{2} &\whitecircle{1} &07/2023   \\
           MemoryBank~\cite{zhong2023memorybank} &- &\whitecircle{2} &\whitecircle{2} &- &\whitecircle{1} &- &07/2023   \\
           %AgentSims\cite{lin2023agentsims} &\whitecircle{1} &\whitecircle{1} &\whitecircle{2} &\whitecircle{2}  &\whitecircle{2} &- &08/2023   \\
           MetaGPT~\cite{hong2023metagpt}& \whitecircle{1} &  \whitecircle{2} & \whitecircle{2} & \whitecircle{2} &\whitecircle{2} &- &08/2023             \\
           % TPTU\cite{ruan2023tptu}& - &  \whitecircle{2} & \whitecircle{1} & \whitecircle{2} &\whitecircle{2} &\whitecircle{3} &08/2023             \\
           \hline \hline
        \end{tabular}\label{construct}
    }
\vspace{-0.5cm}
\end{table*}

\subsection{Agent Capability Acquisition}
In the sections above, we focus mainly on how to design the agent architecture to better harness the capabilities of LLMs to enabling them to accomplish tasks akin to human performance.
The architecture functions as the ``hardware'' of an agent.
However, relying solely on the hardware is insufficient for achieving effective task performance. This is because the agent may lack the necessary task-specific capabilities, skills, and experiences, which can be regarded as "software" resources. In order to equip the agent with these resources, various strategies have been devised.
Generally, we categorize these strategies into two classes based on whether they require fine-tuning of the LLMs.
Below, we introduce each category in detail.

\textbf{Capability Acquisition with Fine-tuning}: 
A direct approach to enhance agent capabilities for task completion is to fine-tune the model using task-specific datasets. These datasets can be constructed from human annotations, LLM-generated content, or real-world applications. We discuss these methods in detail below.

\quad $\bullet$ \textit{Fine-tuning with Human Annotated Datasets}. 
To fine-tune the agent, utilizing human annotated datasets is a versatile approach that can be employed in various application scenarios.
In this approach, researchers first design annotation tasks and then recruit workers to complete them.
For example, in CoH~\cite{liu2023chain}, the authors aim to align LLMs with human values and preferences. Different from the other models, where the human feedback is leveraged in a simple and symbolic manner, this method converts the human feedback into detailed comparison information in the form of natural languages. The LLMs are directly fine-tuned based on these natural language datasets.
In RET-LLM~\cite{modarressi2023ret}, in order to better convert natural languages into structured memory information, the authors fine-tune LLMs based on a human constructed dataset, where each sample is a ``triplet-natural language'' pair.
In WebShop~\cite{yao2022webshop}, the authors collect 1.18 million real-world products from amazon.com, and put them onto a simulated e-commerce website, which contains several carefully designed human shopping scenarios. Based on this website, the authors recruit 13 workers to collect a real-human behavior dataset. At last, three methods based on heuristic rules, imitation learning and reinforcement learning are trained based on this dataset. Although the authors do not fine-tune LLM-based agents, we believe that the dataset proposed in this paper holds immense potential to enhance the capabilities of agents in the field of web shopping.
In EduChat~\cite{dan2023educhat}, the authors aim to enhance the educational functions of LLMs, such as open-domain question answering, essay assessment, Socratic teaching, and emotional support. They fine-tune LLMs based on human annotated datasets that cover various educational scenarios and tasks. 
% These datasets are manually evaluated by psychology experts and frontline teachers.
% SWIFTSAGE~\cite{lin2023swiftsage} is an agent influenced by the dual-process theory of human cognition~\cite{evans2013dual}, which is effective for solving complex interactive reasoning tasks.
% In this agent, the SWIFT module constitutes a compact encoder-decoder language model, which is fine-tuned using human-annotated datasets.

\quad $\bullet$ \textit{Fine-tuning with LLM Generated Datasets}. 
Building human-annotated datasets typically requires recruiting people, which can be costly, especially when dealing with large-scale annotation tasks.
Considering that LLMs can achieve human-like capabilities in a wide range of tasks, a natural idea is using LLMs to accomplish the annotation task.
While the datasets produced from this method can be not as perfect as the human annotated ones, it is much cheaper, and can be leveraged to generate more samples.
For example, in ToolBench~\cite{qin2023toolllm}, to enhance the tool-using capability of open-source LLMs, the authors collect 16,464 real-world APIs spanning 49 categories from the RapidAPI Hub. They used these APIs to prompt ChatGPT to generate diverse instructions, covering both single-tool and multi-tool scenarios. Based on the obtained dataset, the authors fine-tune LLaMA~\cite{touvron2023llama}, and obtain significant performance improvement in terms of tool using.
In~\cite{liu2023training}, to empower the agent with social capability, the authors design a sandbox, and deploy multiple agents to interact with each other.
Given a social question, the central agent first generates initial responses.
Then, it shares the responses to its nearby agents for collecting their feedback.
Based on the feedback as well as its detailed explanations, the central agent revise its initial responses to make them more consistent with social norms.
In this process, the authors collect a large amount of agent social interaction data, which is then leveraged to fine-tune the LLMs.

\quad $\bullet$ \textit{Fine-tuning with Real-world Datasets}.
In addition to building datasets based on human or LLM annotations, directly using real-world datasets to fine-tune the agent is also a common strategy.
For example, in MIND2WEB~\cite{deng2023mind2web}, the authors collect a large amount of real-world datasets to enhance the agent capability in the web domain.
In contrast to prior studies, the dataset presented in this paper encompasses diverse tasks, real-world scenarios, and comprehensive user interaction patterns.
Specifically, the authors collect over 2,000 open-ended tasks from 137 real-world websites spanning 31 domains.
Using this dataset, the authors fine-tune LLMs to enhance their performance on web-related tasks such as movie discovery and ticket booking.
Similarly, in SQL-PaLM~\cite{sun2023sql}, researchers fine-tune PaLM-2 using cross-domain, large-scale text-to-SQL datasets, including Spider and BIRD. The resulting model achieves notable performance improvements on text-to-SQL tasks, demonstrating the effectiveness of real-world datasets for domain-specific applications.

\begin{figure*}[t]
    \centering
    \setlength{\fboxrule}{0.pt}
    \setlength{\fboxsep}{0.pt}
    \fbox{
        \includegraphics[width=1.\linewidth]{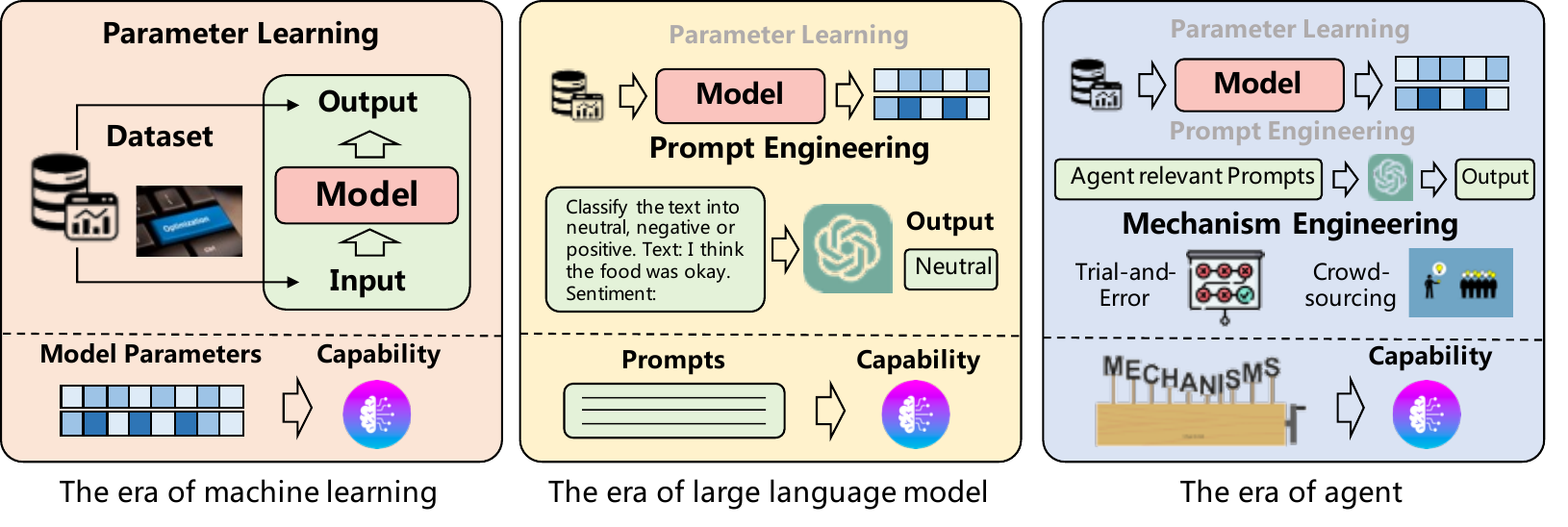}
    }
    \caption{Illustration of transitions in strategies for acquiring model capabilities.}
    \label{cap}
    \vspace{-0.cm}
\end{figure*}

\textbf{Capability Acquisition without Fine-tuning}:
In the era of tradition machine learning, the model capability is mainly acquired by learning from datasets, where the knowledge is encoded into the model parameters.
In the era of LLMs, the model capability can be acquired either by training/fine-tuning the model parameters or designing delicate prompts (\emph{i.e.}, prompt engineering).
In prompt engineering, one needs to write valuable information into the prompts to enhance the model capability or unleash existing LLM capabilities.
In the era of agents, the model capability can be acquired based on three strategies: 
(1) model fine-tuning, (2) prompt engineering and (3) designing proper agent evolution mechanisms (we called it as \textit{mechanism engineering}).
Mechanism engineering is a broad concept that involves developing specialized modules, introducing novel working rules, and other strategies to enhance agent capabilities.
{For clearly understanding the transitions of model capability acquisition strategies, we illustrate them in Figure~\ref{cap}.}
% In the above section, we have detailed the strategy of fine-tuning.
In the following, we detail prompting engineering and mechanism engineering.

\quad $\bullet$ \textit{Prompting Engineering}. 
Due to the strong language comprehension capabilities, people can directly interact with LLMs using natural languages.
This introduces a novel strategy for enhancing agent capabilities, that is, one can describe the desired capability using natural language and then use it as prompts to influence LLM actions.
For example, in CoT~\cite{wei2022chain}, in order to empower the agent with the capability for complex task reasoning, the authors present the intermediate reasoning steps as few-shot examples in the prompt.
Similar techniques are also used in CoT-SC~\cite{wang2022self} and ToT~\cite{yao2023tree}.
In RLP~\cite{fischer2023reflective}, the authors aim to enhance an agent’s self-awareness in conversations by prompting LLMs with the agent’s beliefs about both its own and the listeners’ mental states. This approach results in more engaging and adaptive utterances. Furthermore, the incorporation of the target mental states of listeners allows the agent to formulate more strategic plans.
Retroformer~\cite{yao2023retroformer} presents a retrospective model that enables the agent to generate reflections on its past failures.
The reflections are integrated into the prompt of LLMs to guide the agent's future actions.
Additionally, this model utilizes reinforcement learning to iteratively improve the retrospective model, thereby refining the LLM prompt.

\quad $\bullet$ \textit{Mechanism Engineering}. 
Unlike model fine-tuning and prompt engineering, mechanism engineering is a unique strategy to enhance agent capability.
In the following, we present several representative methods of mechanism engineering.

(1) \textit{Trial-and-error.}
In this method, the agent first performs an action, and subsequently, a pre-defined critic is invoked to judge the action.
If the action is deemed unsatisfactory, then the agent reacts by incorporating the critic's feedback.
For example, in RAH~\cite{shu2023rah}, the agent serves as a user assistant in recommender systems. One of the agent's crucial roles is to simulate human behavior and generate responses on behalf of the user. To fulfill this objective, the agent first generates a predicted response and then compares it with the real human feedback. If the predicted response and the real human feedback differ, the critic generates failure information, which is subsequently incorporated into the agent's next action.
Similarly, in DEPS~\cite{wang2023describe}, the agent first designs a plan to accomplish a given task.
In the plan execution process, if an action fails, the explainer generates specific details explaining the cause of the failure. This information is then incorporated by the agent to redesign the plan.
In RoCo~\cite{mandi2023roco}, the agent first proposes a sub-task plan and a path of 3D waypoints for each robot in a multi-robot collaboration task. 
The plan and waypoints are then validated by a set of environment checks, such as collision detection and inverse kinematics. If any of the checks fail, the feedback is appended to each agent's prompt and another round of dialog begins. The agents use LLMs to discuss and improve their plan and waypoints until they pass all validations.
PREFER~\cite{zhang2023prefer} extends this idea by leveraging LLMs to generate detailed feedback when the agent underperforms, enabling iterative refinement and performance improvement.

(2) \textit{Crowd-sourcing.} In~\cite{du2023improving}, the authors design a debating mechanism that leverages the wisdom of crowds to enhance agent capabilities.
To begin with, different agents provide separate responses to a given question.
If their responses are not consistent, they will be prompted to incorporate the solutions from other agents and provide an updated response.
This iterative process continues until reaching a final consensus answer.
In this method, the capability of each agent is enhanced by understanding and incorporating the other agents' opinions.

(3) \textit{Experience Accumulation.}
In GITM~\cite{zhu2023ghost}, the agent does not know how to solve a task in the beginning.
Then, it makes explorations, and once it has successfully accomplished a task, the actions used in this task are stored into the agent memory.
In the future, if the agent encounters a similar task, then the relevant memories are extracted to complete the current task.
In this process, the improved agent capability comes from the specially designed memory accumulation and utilization mechanisms.
Voyager~\cite{wang2023voyager} introduces a skill library, where executable codes for specific skills are refined through interactions with the environment, enabling efficient task execution over time. 
In AppAgent~\cite{yang2023appagent}, the agent is designed to interact with apps in a manner akin to human users, learning through both autonomous exploration and observation of human demonstrations. Throughout this process, it constructs a knowledge base, which serves as a reference for performing intricate tasks across various applications on a mobile phone.
In MemPrompt~\cite{madaan2022memory}, the users are requested to provide feedback in natural language regarding the problem-solving intentions of the agent, and this feedback is stored in memory. When the agent encounters similar tasks, it attempts to retrieve related memories to generate more suitable responses.

(4) \textit{Self-driven Evolution.}
This method allows agents to autonomously improve through self-directed learning and feedback mechanisms. 
LMA3~\cite{colas2023augmenting} enables the agent to autonomously set goals for itself, and gradually improve its capability by exploring the environment and receiving feedback from a reward function.
Following this mechanism, the agent can acquire knowledge and develop capabilities according to its own preferences.
SALLM-MS~\cite{nascimento2023self} integrates advanced LLMs like GPT-4 into a multi-agent system, agents can adapt and perform complex tasks, showcasing advanced communication capabilities, thereby realizing self-driven evolution in their interactions with the environment.
In CLMTWA~\cite{saha2023can}, by using a large language model as a teacher and a weaker language model as a student, the teacher can generate and communicate natural language explanations to improve the student’s reasoning skills via theory of mind. The teacher can also personalize its explanations for the student and intervene only when necessary, based on the expected utility of intervention. 
Meanwhile, NLSOM~\cite{zhuge2023mindstorms},leverages natural language collaboration between agents, dynamically adjusting roles, tasks, and relationships based on feedback to solve problems beyond the scope of a single agent.

\begin{remark}
Upon comparing the aforementioned strategies for agent capability acquisition, we can find that the fine-tuning method improves the agent capability by adjusting model parameters, which can incorporate a large amount of task-specific knowledge, but is only suitable for open-source LLMs. The method without fine-tuning usually enhances the agent capability based on delicate prompting strategies or mechanism engineering. They can be used for both open- and closed-source LLMs. However, due to the limitation of the input context window of LLMs, they cannot incorporate too much task information.
In addition, the designing spaces of the prompts and mechanisms are extremely large, which makes it not easy to find optimal solutions.
\end{remark}

In the above sections, we have detailed the construction of LLM-based agents, where we focus on two aspects including the architecture design and capability acquisition. We present the correspondence between existing work and the above taxonomy in Table~\ref{construct}. 
It should be noted that, for the sake of integrity, we have also incorporated several studies, which do not explicitly mention LLM-based agents but are highly related to this area.

\begin{figure*}[t]
    \centering
    \setlength{\fboxrule}{0.pt}
    \setlength{\fboxsep}{0.pt}
    \fbox{
        \includegraphics[width=1.\linewidth]{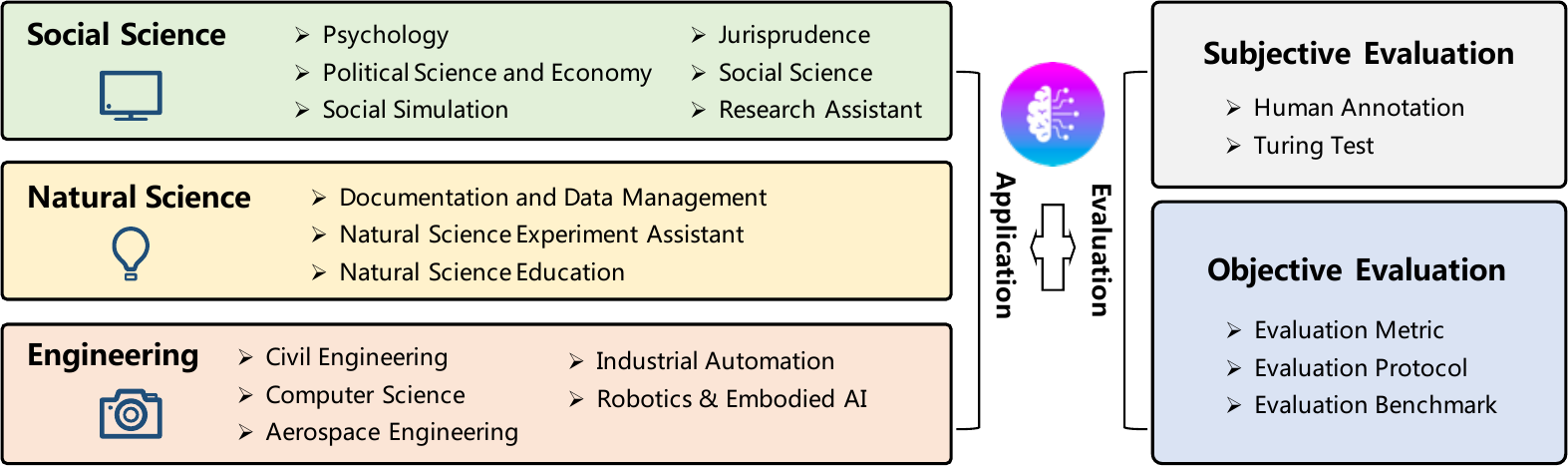}
    }
    \caption{The applications (left) and evaluation strategies (right) of LLM-based agents.}
    \label{ae}
    \vspace{-0.cm}
\end{figure*}

\section{LLM-based Autonomous Agent Application}\label{sec:application}
Owing to the strong language comprehension, complex task reasoning, and common sense understanding capabilities, LLM-based autonomous agents have shown significant potential to influence multiple domains.
This section provides a succinct summary of previous studies, categorizing them according to their applications in three distinct areas: social science, natural science, and engineering (see the left part of Figure~\ref{ae} for a global overview). 

\subsection{Social Science}
Social science is one of the branches of science, devoted to the study of societies and the relationships among individuals within those societies.
% \footnote{https://en.wikipedia.org/wiki/Social\_science}.
LLM-based autonomous agents can promote this domain by leveraging their impressive human-like understanding, thinking and task solving capabilities.
In the following, we discuss several key areas that can be affected by LLM-based autonomous agents.

\textbf{Psychology}: 
For the domain of psychology, LLM-based agents can be leveraged for conducting simulation experiments, providing mental health support and so on~\cite{aher2023using,akata2023playing,ma2023understanding,ziems2023can}.	
For example, in~\cite{aher2023using}, the authors assign LLMs with different profiles, and let them complete psychology experiments.
From the results, the authors find that LLMs are capable of generating results that align with those from studies involving human participants.
% In addition, larger models can usually provide more faithful simulation results than the smaller ones.
Additionally, it was observed that larger models tend to deliver more accurate simulation results compared to their smaller counterparts.
An interesting discovery is that, in many experiments, models like ChatGPT and GPT-4 can provide too perfect estimates (called ``hyper-accuracy distortion''), which may influence the downstream applications.
In~\cite{ma2023understanding}, the authors systematically analyze the effectiveness of LLM-based conversation agents for mental well-being support.
They collect 120 posts from Reddit, and find that such agents can help users cope with anxieties, social isolation and depression on demand.
At the same time, they also find that the agents may produce harmful contents sometimes.

\textbf{Political Science and Economy}: 
LLM-based agents can also be utilized to study political science and economy~\cite{argyle2023out,ziems2023can,horton2023large}. 
In~\cite{argyle2023out}, LLM-based agents are utilized for ideology detection and predicting voting patterns.
In~\cite{ziems2023can}, the authors focuses on understanding the discourse structure and persuasive elements of political speech through the assistance of LLM-based agents.
In~\cite{horton2023large}, LLM-based agents are provided with specific traits such as talents, preferences, and personalities to explore human economic behaviors in simulated scenarios.

\textbf{Social Simulation}: 
Previously, conducting experiments with human societies is often expensive, unethical, or even infeasible. 
With the ever prospering of LLMs, many people explore to build virtual environment with LLM-based agents to simulate social phenomena, such as the propagation of harmful information, and so on~\cite{park2022social,li2023you,li2023quantifying,park2023generative,lin2023agentsims,kovavc2023socialai,gao2023s,williams2023epidemic}. 
For example, Social Simulacra~\cite{park2022social} simulates an online social community and explores the potential of utilizing agent-based simulations to aid decision-makers to improve community regulations. 
~\cite{li2023you, li2023quantifying} investigates the potential impacts of different behavioral characteristics of LLM-based agents in social networks.
Generative Agents~\cite{park2023generative} and AgentSims\cite{lin2023agentsims} construct multiple agents in a virtual town to simulate the human daily life. 
SocialAI School~\cite{kovavc2023socialai} employs LLM-based agents to simulate and investigate the fundamental social cognitive skills during the course of child development.
{S$^3$}~\cite{gao2023s} builds a social network simulator, focusing on the propagation of information, emotion and attitude.
{CGMI~\cite{jinxin2023cgmi} is a framework for multi-agent simulation. CGMI maintains the personality of the agents through a tree structure and constructs a cognitive model. The authors simulated a classroom scenario using CGMI.}

\textbf{Jurisprudence}: LLM-based agents can serve as aids in legal decision-making processes, facilitating more informed judgements~\cite{cui2023chatlaw,hamilton2023blind}. 
Blind Judgement~\cite{hamilton2023blind} employs several language models to simulate the decision-making processes of multiple judges. It gathers diverse opinions and consolidates the outcomes through a voting mechanism. ChatLaw~\cite{cui2023chatlaw} is a prominent Chinese legal model based on LLM. It adeptly supports both database and keyword search strategies, specifically designed to mitigate the hallucination issue prevalent in such models.
In addition, this model also employs self-attention mechanism to enhance the LLM's capability via mitigating the impact of reference inaccuracies.

\textbf{Research Assistant}: 
% Beyond their application in specialized domains, LLM-based agents are increasingly adopted as versatile assistants in the broad field of social science research~\cite{bail2023can,ziems2023can}. 
% In~\cite{ziems2023can}, LLM-based agents offer multifaceted assistance, ranging from generating concise article abstracts and extracting pivotal keywords to crafting detailed scripts for studies.
% In~\cite{bail2023can}, LLM-based agents serve as a writing assistant, where they possess the capability to identify novel research inquiries for social scientists.
Beyond their application in specialized domains, LLM-based agents are increasingly adopted as versatile assistants in the broad field of social science research~\cite{bail2023can,ziems2023can}. In~\cite{ziems2023can}, LLM-based agents offer multifaceted assistance, ranging from generating concise article abstracts and extracting pivotal keywords to crafting detailed scripts for studies, showcasing their ability to enrich and streamline the research process. Meanwhile, in~\cite{bail2023can}, LLM-based agents serve as a writing assistant, demonstrating their capability to identify novel research inquiries for social scientists, thereby opening new avenues for exploration and innovation in the field. These examples highlight the  potential of LLM-based agents in enhancing the efficiency, creativity, and breadth of social science research.

\subsection{Natural Science}
Natural science is one of the branches of science concerned with the description, understanding and prediction of natural phenomena, based on empirical evidence from observation and experimentation.
% \footnote{https://en.wikipedia.org/wiki/Natural\_science}.
With the ever prospering of LLMs, the application of LLM-based agents in natural sciences becomes more and more popular.
In the following, we present many representative areas, where LLM-based agents can play important roles.

\textbf{Documentation and Data Management}: 
Natural scientific research often involves the collection, organization, and synthesis of substantial amounts of literature, which requires a significant dedication of time and human resources.
LLM-based agents have shown strong capabilities on language understanding and employing tools such as the internet and databases for text processing.
These capabilities empower the agent to excel in tasks related to documentation and data management.
In~\cite{boiko2023emergent}, the agent can efficiently query and utilize internet information to complete tasks such as question answering and experiment planning. 
ChatMOF~\cite{kang2023chatmof} utilizes LLMs to extract important information from text descriptions written by humans. 
It then formulates a plan to apply relevant tools for predicting the properties and structures of metal-organic frameworks.
ChemCrow~\cite{bran2023chemcrow} utilizes chemistry-related databases to both validate the precision of compound representations and identify potentially dangerous substances. This functionality enhances the reliability and comprehensiveness of scientific inquiries by ensuring the accuracy of the data involved.

\textbf{Experiment Assistant}:
LLM-based agents have the ability to independently conduct experiments, making them valuable tools for supporting scientists in their research projects~\cite{boiko2023emergent,bran2023chemcrow}.
For instance, ~\cite{boiko2023emergent} introduces an innovative agent system that utilizes LLMs for automating the design, planning, and execution of scientific experiments. 
This system, when provided with the experimental objectives as input, accesses the Internet and retrieves relevant documents to gather the necessary information. 
It subsequently utilizes Python code to conduct essential calculations and carry out the following experiments.
ChemCrow~\cite{bran2023chemcrow} incorporates 17 carefully developed tools that are specifically designed to assist researchers in their chemical research. Once the input objectives are received, ChemCrow provides valuable recommendations for experimental procedures, while also emphasizing any potential safety risks associated with the proposed experiments.

\textbf{Natural Science Education}:
LLM-based agents can communicate with humans fluently, often being utilized to develop agent-based educational tools.
For example,~\cite{boiko2023emergent} develops agent-based education systems to facilitate students learning of experimental design, methodologies, and analysis. 
The objective of these systems is to enhance students' critical thinking and problem-solving skills, while also fostering a deeper comprehension of scientific principles.
Math Agents~\cite{swan2023math} can assist researchers in exploring, discovering, solving and proving mathematical problems. 
Additionally, it can communicate with humans and aids them in understanding and using mathematics.
~\cite{Drori_2022} utilize the capabilities of CodeX~\cite{chen2021evaluating} to automatically solve and explain university-level mathematical problems, which can be used as education tools to teach students and researchers.
CodeHelp~\cite{liffiton2023codehelp} is an education agent for programming. It offers many useful features, such as setting course-specific keywords, monitoring student queries, and providing feedback to the system.
EduChat~\cite{dan2023educhat} is an LLM-based agent designed specifically for the education domain. 
It provides personalized, equitable, and empathetic educational support to teachers, students, and parents through dialogue.
FreeText~\cite{matelsky2023large} is an agent that utilizes LLMs to automatically assess students' responses to open-ended questions and offer feedback.

\setlength{\textfloatsep}{0.4cm}

\begin{table*}[ht!]
\centering
\caption{Representative applications of LLM-based autonomous agents.}
\renewcommand\arraystretch{1.6}
\scalebox{0.9}{
\begin{tabular}{m{4cm}|m{5cm}|m{8.8cm}}
\hline\hline
 & Domain & Work \\
\hline
 & Psychology & TE~\cite{aher2023using}, Akata et al.~\cite{akata2023playing}, Ziems et al.~\cite{ziems2023can}, Ma et al.~\cite{ma2023understanding} \\\cline{2-3}
\multirow{5}{*}{\parbox{2cm}{Social Science}} 
 & Political Science and Economy & Argyle et al.~\cite{argyle2023out}, Horton~\cite{horton2023large}, Ziems et al.~\cite{ziems2023can} \\\cline{2-3}
 & Social Simulation & Social Simulacra~\cite{park2022social}, Generative Agents~\cite{park2023generative}, SocialAI School~\cite{kovavc2023socialai}, AgentSims~\cite{lin2023agentsims}, S$^3$~\cite{gao2023s}, Williams et al.~\cite{williams2023epidemic}, Li et al.~\cite{li2023you}, Chao et al.~\cite{li2023quantifying}\\\cline{2-3}
 & Jurisprudence & ChatLaw~\cite{cui2023chatlaw}, Blind Judgement~\cite{hamilton2023blind} \\\cline{2-3}
 & Research Assistant & Ziems et al.~\cite{ziems2023can}, Bail et al.~\cite{bail2023can} \\
\hline
 & \parbox{4cm}{Documentation and \\Data Management} & ChemCrow~\cite{bran2023chemcrow}, ChatMOF~\cite{kang2023chatmof}, Boiko et al.~\cite{boiko2023emergent} \\\cline{2-3}
\multirow{3}{*}[20pt]{\parbox{3cm}{Natural Science}}
 & Experiment Assistant & ChemCrow~\cite{bran2023chemcrow}, Boiko et al.~\cite{boiko2023emergent}, Grossmann et al.~\cite{grossmann2023ai} \\\cline{2-3}
 & Natural Science Education & ChemCrow~\cite{bran2023chemcrow}, CodeHelp~\cite{liffiton2023codehelp}, Boiko et al.~\cite{boiko2023emergent}, MathAgent~\cite{swan2023math}, Drori et al.~\cite{Drori_2022}, EduChat~\cite{dan2023educhat}, FreeText~\cite{matelsky2023large}\\
\hline
\multirow{3}{*}[-23pt]{Engineering}
 & CS \& SE & RestGPT~\cite{song2023restgpt}, Self-collaboration~\cite{dong2023self}, SQL-PALM~\cite{sun2023sql}, RAH~\cite{shu2023rah}, D-Bot~\cite{zhou2023llm}, RecMind~\cite{wang2023recmind}, ChatEDA~\cite{he2023chateda}, InteRecAgent~\cite{huang2023recommender}, PentestGPT~\cite{deng2023pentestgpt}, CodeHelp~\cite{liffiton2023codehelp}, SmolModels~\cite{SmolModels}, DemoGPT~\cite{DemoGPT}, GPTEngineer~\cite{GPTEngineer} \\\cline{2-3}
 & Industrial Automation & GPT4IA~\cite{xia2023towards}, IELLM~\cite{ogundare2023industrial}\\\cline{2-3}
 & Robotics \& Embodied AI & ProAgent~\cite{zhang2023proagent}, LLM4RL~\cite{huenabling}, PET~\cite{wu2023plan}, REMEMBERER~\cite{zhang2023large}, DEPS~\cite{wang2023describe}, Unified Agent~\cite{di2023towards}, SayCan~\cite{ahn2022can}, TidyBot~\cite{wu2023tidybot}, RoCo~\cite{mandi2023roco}, SayPlan~\cite{rana2023sayplan}, TaPA~\cite{wu2023embodied}, 
 Dasgupta et al.~\cite{dasgupta2023collaborating}, DECKARD~\cite{nottingham2023embodied}, Dialogue shaping~\cite{zhou2023dialogue}\\
\hline\hline
\end{tabular}
}
\label{application}
\end{table*}

\subsection{Engineering}

LLM-based autonomous agents have shown great potential in assisting and enhancing engineering research and applications. In this section, we review and summarize the applications of LLM-based agents in several major engineering domains.

\textbf{Computer Science \& Software Engineering}:
In the field of computer science and software engineering, LLM-based agents offer potential for automating coding, testing, debugging, and documentation generation~\cite{qian2023communicative,hong2023metagpt,dong2023self,GPTEngineer,SmolModels,DemoGPT}. ChatDev~\cite{qian2023communicative} proposes an end-to-end framework, where multiple agent roles communicate and collaborate through natural language conversations to complete the software development life cycle. This framework demonstrates efficient and cost-effective generation of executable software systems. 
MetaGPT~\cite{hong2023metagpt} abstracts multiple roles, such as product managers, architects, project managers, and engineers, to supervise code generation process and enhance the quality of the final output code. This enables low-cost software development. 
~\cite{dong2023self} presents a self-collaboration framework for code generation using LLMs. 
In this framework, multiple LLMs are assumed to be distinct "experts" for specific sub-tasks. 
They collaborate and interact according to specified instructions, forming a virtual team that facilitates each other's work. 
Ultimately, the virtual team collaboratively addresses code generation tasks without requiring human intervention. 
% LLMs can also be leveraged to detect and correct code bugs. 
LLIFT~\cite{li2023hitchhiker} employs LLMs to assist in conducting static analysis, specifically for identifying potential code vulnerabilities. This approach effectively manages the trade-off between accuracy and scalability.
ChatEDA~\cite{he2023chateda} is an agent developed for electronic design automation (EDA) to streamline the design process by integrating task planning, script generation, and execution.
CodeHelp~\cite{liffiton2023codehelp} is an agent designed to assist students and developers in debugging and testing their code. Its features include providing detailed explanations of error messages, suggesting potential fixes, and ensuring the accuracy of the code.
Pentest~\cite{deng2023pentestgpt} is a penetration testing tool based on LLMs, which can effectively identify common vulnerabilities, and interpret source code to develop exploits.
D-Bot~\cite{zhou2023llm} utilizes the capabilities of LLMs to systematically assess potential root causes of anomalies in databases. 
Through the implementation of a tree of thought approach, D-Bot enables LLMs to backtrack to previous steps in case the current step proves unsuccessful, thus enhancing the accuracy of the diagnosis process.

\textbf{Industrial Automation}: In the field of industrial automation, LLM-based agents can be used to achieve intelligent planning and control of production processes. \cite{xia2023towards} proposes a novel framework that integrates LLMs with digital twin systems to accommodate flexible production needs. The framework leverages prompt engineering techniques to create LLM agents that can adapt to specific tasks based on the information provided by digital twins. These agents can coordinate a series of atomic functionalities and skills to complete production tasks at different levels. This research demonstrates the potential of integrating LLMs into industrial automation systems, providing innovative solutions for more agile, flexible and adaptive production processes.
% IELLM~\cite{ogundare2023industrial} is a case study of ChatGPT’s performance on oil and gas problems, such as rock physics modeling, acoustic reflectometry, and coiled tubing control, which leverages LLM-based autonomous agent to generate mathematical models, solve partial differential equations, and extrapolate the theory to different scenarios.
% IELLM~\cite{ogundare2023industrial} presents a comprehensive case study on LLMs' effectiveness in addressing challenges in the oil and gas industry. 
% It focuses on various applications, including rock physics modeling, acoustic reflectometry, and coiled tubing control. 
IELLM~\cite{ogundare2023industrial} showcases a case study on LLMs' role in the oil and gas industry, covering applications like factory automation and PLC programming.

% The framework employs an LLM-based autonomous agent, which successfully generates mathematical models, solves partial differential equations, and extrapolates theoretical concepts to different scenarios.

\textbf{Robotics \& Embodied Artificial Intelligence}:
Recent works have advanced the development of more efficient reinforcement learning agents for robotics and embodied artificial intelligence~\cite{dasgupta2023collaborating,zhou2023dialogue,nottingham2023embodied,wu2023embodied,wang2023voyager,zhu2023ghost,huenabling,wu2023plan,zhang2023large,di2023towards,ahn2022can}. These efforts focus on enhancing autonomous agents’ capabilities in planning, reasoning, and collaboration within embodied environments.
For instance,~\cite{dasgupta2023collaborating} proposes the Planner-Actor-Reporter paradigm for embodied reasoning and task planning. 
% In this system, the authors design high-level commands to enable improved planning while propose low-level controllers to translate commands into actions. 
% Additionally, one can leverage dialogues to gather information~\cite{zhou2023dialogue} to accelerate the optimization process.
% ~\cite{zhou2023dialogue} features a dialogue module that interacts with NPCs to gather key information, which is then used to create knowledge graphs that guide the RL agent, optimizing the decision-making process.
% Similarly, ~\cite{nottingham2023embodied,wu2023embodied} adopt distinct mechanisms to enable autonomous agents for embodied decision-making and exploration, further enhancing the agents' performance in complex environments.
DECKARD~\cite{nottingham2023embodied} introduces the Planner-Actor-Reporter paradigm, which facilitates embodied reasoning and task planning by decoupling the agent’s planning, execution, and reporting processes.
TaPA~\cite{wu2023embodied} constructs a multimodal dataset comprising multi-view RGB images of indoor scenes, human instructions, and corresponding plans to fine-tune LLMs. The fine-tuned models align visual perception with task planning, enabling them to generate more executable plans and significantly improving their performance in visually grounded tasks.
% ~\cite{zhou2023dialogue} features a dialogue module that interacts with NPCs to gather key information, which is then used to create knowledge graphs that guide the RL agent, optimizing the decision-making process. Similarly, ~\cite{nottingham2023embodied,wu2023embodied} employ autonomous agents for embodied decision-making and exploration, further enhancing the agents' performance in complex environments.

To overcome the physical constraints, the agents can generate executable plans and accomplish long-term tasks by leveraging multiple skills.
In terms of control policies, SayCan~\cite{ahn2022can} focuses on investigating a wide range of manipulation and navigation skills utilizing a mobile manipulator robot. Taking inspiration from typical tasks encountered in a kitchen environment, it presents a comprehensive set of 551 skills that cover seven skill families and 17 objects. These skills encompass various actions such as picking, placing, grasping, and manipulating objects, among others.
% Additional frameworks such as VOYAGAR~\cite{wang2023voyager} and GITM~\cite{zhu2023ghost} propose autonomous agents that communicate, collaborate, and accomplish complex tasks. This demonstrates the promise of natural language understanding, motion planning, and human interaction for real-world robotics. As capabilities advance, adaptive autonomous agents may accomplish increasingly complex embodied tasks.
% At last, complementing conventional methods with reasoning and planning abilities as in~\cite{huenabling,wu2023plan,zhang2023large,di2023towards} can significantly improve autonomous agent performance in embodied environments. The focus is on holistic systems that enhance sample efficiency, generalization, and accomplish long-horizon tasks.
TidyBot~\cite{wu2023tidybot} is an embodied agent designed to personalize household cleanup tasks.
It can learn users' preferences on object placement and manipulation methods through textual examples.

To promote the application of LLM-based autonomous agents, researchers have also introduced many open-source libraries, based on which the developers can quickly implement and evaluate agents according to their customized requirements~\cite{Auto-gpt,Agentgpt,AI-legion,AGiXT,AgentVerse,XLang,BabyAGI,langchain,workgpt,loopgpt,gpt-researcher,qin2023bmtools,transformers-agent,DemoGPT,MiniAGI,SuperAGI,wu2023autogen}.
For example, 
LangChain~\cite{langchain} is an open-source framework that automates coding, testing, debugging, and documentation generation tasks. By integrating language models with data sources and facilitating interaction with the environment, LangChain enables efficient and cost-effective software development through natural language communication and collaboration among multiple agent roles.
Based on LangChain, XLang~\cite{XLang} provides a comprehensive set of tools and a fully integrated user interface. It focuses on executable language grounding, enabling the conversion of natural language instructions into code or action sequences that interact seamlessly with various environments, including databases, web applications, and physical robots.
AutoGPT~\cite{Auto-gpt} is an agent that is fully automated. It sets one or more goals, breaks them down into corresponding tasks, and cycles through the tasks until the goal is achieved. 
WorkGPT~\cite{workgpt} is an agent framework similar to AutoGPT and LangChain. By providing it with an instruction and a set of APIs, it engages in back-and-forth conversations with AI until the instruction is completed. 
% LoopGPT~\cite{loopgpt} is a modular Auto-GPT framework that re-implements the popular Auto-GPT project as a proper Python package, designed with modularity and extensibility in mind.
GPT-Engineer~\cite{GPTEngineer} and DemoGPT~\cite{DemoGPT} are open-source projects that focus on automating code generation through prompts to complete development tasks.
SmolModels~\cite{SmolModels} offers a family of compact language models suitable for various tasks.
AGiXT~\cite{AGiXT} is a dynamic AI automation platform that efficiently manages instructions and executes complex tasks across various AI providers, integrating adaptive memory, smart features, and a versatile plugin system.
AgentVerse~\cite{chen2023agentverse} is a versatile framework that facilitates researchers in creating customized LLM-based agent simulations efficiently.
GPT Researcher~\cite{gpt-researcher} is an experimental application that leverages LLMs to efficiently develop research questions, trigger web crawls to gather information, summarize sources, and aggregate summaries.
BMTools~\cite{qin2023bmtools} provides a platform for community-driven tool building and sharing. It supports various types of tools, enables simultaneous task execution using multiple tools, and offers a simple interface for loading plugins via URLs, fostering easy development and contribution to the BMTools ecosystem. 

\begin{remark}
Utilization of LLM-based agents in supporting above applications may also entail risks and challenges. 
On one hand, LLMs themselves may be susceptible to illusions and other issues, occasionally providing erroneous answers, leading to incorrect conclusions, experimental failures, or even posing risks to human safety in hazardous experiments. Therefore, during experimentation, users must possess the necessary expertise and knowledge to exercise appropriate caution. On the other hand, LLM-based agents could potentially be exploited for malicious purposes, such as the development of chemical weapons, necessitating the implementation of security measures, such as human alignment, to ensure responsible and ethical use.

In summary, in the above sections, we introduce the typical applications of LLM-based autonomous agents in three important domains. To facilitate a clearer understanding, we have summarized the relationship between previous studies and their respective applications in Table~\ref{application}.
\end{remark}

\section{LLM-based Autonomous Agent Evaluation}\label{sec:evaluation}
Similar to LLMs themselves, evaluating the effectiveness of LLM-based autonomous agents is a challenging task.
% This section introduces two commonly used evaluation strategies, that is, subjective and objective evaluation (see the right part of Figure~\ref{ae} for an overview).
This section outlines two prevalent approaches to evaluation: subjective and objective methods. For a comprehensive overview, please refer to the right portion of Figure~\ref{ae}.

\subsection{Subjective Evaluation}
Subjective evaluation measures the agent capabilities based on human judgements~\cite{lee2022evaluating,park2022social,park2023generative,argyle2023out,zhang2023building}. 
It is suitable for the scenarios where there are no evaluation datasets or it is very hard to design quantitative metrics, for example, evaluating the agent's intelligence or user-friendliness.
In the following, we present two commonly used strategies for subjective evaluation.

\textbf{Human Annotation}: 
This evaluation method involves human evaluators directly scoring or ranking the outputs generated by various agents~\cite{ziems2023can,argyle2023out,zhang2023building}.
%For example, in~\cite{park2023generative}, the authors employ many annotators, and ask them to provide feedback on five key questions that directly associated with the agent capability.
For example, in~\cite{park2023generative}, the authors engage numerous annotators by asking 25 questions that explore their abilities across five key areas directly related to agent capabilities.
In~\cite{park2022social}, annotators are asked to determine whether the specifically designed models can significantly enhance the development of rules within online communities.

\textbf{Turing Test}: 
This evaluation strategy necessitates that human evaluators differentiate between outputs produced by agents and those created by humans.
If, in a given task, the evaluators cannot separate the agent and human results, it demonstrates that the agent can achieve human-like performance on this task.
For instance, researchers in~\cite{argyle2023out} conduct experiments on free-form Partisan text, and the human evaluators are asked to guess whether the responses are from human or LLM-based agent.

\begin{remark}
LLM-based agents are usually designed to serve humans. 
Thus, subjective agent evaluation plays a critical role, since it reflects human criterion.
However, this strategy also faces issues such as high costs, inefficiency, and population bias. 
To address these issues, a growing number of researchers are investigating the use of LLMs themselves as intermediaries for carrying out these subjective assessments.
For example, in ChemCrow~\cite{bran2023chemcrow}, researchers assess the experimental results using GPT. They consider both the completion of tasks and the accuracy of the underlying processes.
Similarly, ChatEval~\cite{chan2023chateval} introduces a novel approach by employing multiple agents to critique and assess the results generated by various candidate models in a structured debate format.
This innovative use of LLMs for evaluation purposes holds promise for enhancing both the credibility and applicability of subjective assessments in the future. As LLM technology continues to evolve, it is anticipated that these methods will become increasingly reliable and find broader applications, thereby overcoming the current limitations of direct human evaluation.
\end{remark}

\begin{table*}[ht!]
    \centering
    \caption{
    % Summary on the evaluation strategies of LLM-based autonomous agents (more agents can be seen on https://github.com/Paitesanshi/LLM-Agent-Survey). 
    For subjective evaluation, we use \whitecircle{1} and \whitecircle{2} to represent human annotation and the Turing test, respectively.
    For objective evaluation, we use \whitecircle{1}, \whitecircle{2}, \whitecircle{3}, and \whitecircle{4} to represent real-world simulation, social evaluation, multi-task evaluation, and software testing, respectively. ``$\checkmark$'' indicates that the evaluations are based on benchmarks.}
    % \vspace{0.2cm}
    \renewcommand\arraystretch{0.9}
    \scalebox{.93}{
        \begin{tabular}{p{4.1cm}p{3.1cm}<{\centering}p{3.1cm}<{\centering}p{3.1cm}<{\centering}p{3.1cm}<{\centering}}
            \hline\hline
            {Model} & {Subjective } & {Objective } &Benchmark&Time\\
            % \cline{2-4}
            \hline
            WebShop~\cite{yao2022webshop} & - & \whitecircle{1} \whitecircle{3}& $\checkmark$ & 07/2022 \\            
            Social Simulacra~\cite{park2022social} & \whitecircle{1} & \whitecircle{2} & - & 08/2022 \\
            TE~\cite{aher2023using} & - & \whitecircle{2} & - & 08/2022 \\
            LIBRO~\cite{kang2023large} & - & \whitecircle{4} & - & 09/2022 \\
            ReAct~\cite{yao2022react} & - & \whitecircle{1} & $\checkmark$ & 10/2022 \\
            Argyle et al.~\cite{argyle2023out} & \whitecircle{2} & \whitecircle{2} \whitecircle{3} & - & 02/2023 \\
            DEPS~\cite{wang2023describe} & - & \whitecircle{1} & $\checkmark$ & 02/2023 \\
            Jalil et al.~\cite{jalil2023chatgpt} & - & \whitecircle{4} & - & 02/2023\\
            Reflexion~\cite{shinn2023reflexion} & - & \whitecircle{1} \whitecircle{3} & - & 03/2023 \\
            IGLU~\cite{mehta2023improving} & - & \whitecircle{1} & $\checkmark$ & 04/2023 \\
            %LLM+P\cite{liu2023llmp+} & - & - & - & 04/2023 \\
            Generative Agents~\cite{park2023generative} & \whitecircle{1}  & - & - & 04/2023 \\
            ToolBench~\cite{qin2023bmtools} & - & \whitecircle{3} & $\checkmark$ & 04/2023 \\
            GITM~\cite{zhu2023ghost} & - & \whitecircle{1} & $\checkmark$ & 05/2023 \\
            Two-Failures~\cite{chen2023two} & - &\whitecircle{3} &- & 05/2023 \\
            Voyager~\cite{wang2023voyager} & - & \whitecircle{1} & $\checkmark$ & 05/2023 \\
            SocKET~\cite{choi2023socket} & - & \whitecircle{2} \whitecircle{3} & $\checkmark$ & 05/2023 \\
            MobileEnv~\cite{zhang2023mobileenv} & - & \whitecircle{1} \whitecircle{3} & $\checkmark$ & 05/2023 \\
            Clembench~\cite{chalamalasetti2023clembench} & - & \whitecircle{1} \whitecircle{3} & $\checkmark$ & 05/2023 \\
            Dialop~\cite{lin2023decision} & - & \whitecircle{3} & $\checkmark$ & 06/2023 \\
            Feldt et al.~\cite{feldt2023towards} & - & \whitecircle{4} & - & 06/2023 \\
            CO-LLM~\cite{zhang2023building} & \whitecircle{1} & \whitecircle{1} & - & 07/2023 \\
            Tachikuma~\cite{liang2023tachikuma} & \whitecircle{1} & \whitecircle{1} \whitecircle{3} & $\checkmark$ & 07/2023 \\
            %ChatDev\cite{qian2023communicative} & - &  & - & 07/2023 \\
            %WebArena~\cite{zhou2023webarena}& - & \whitecircle{1}  & $\checkmark$ &  07/2023 \\
            RocoBench~\cite{mandi2023roco} & - & \whitecircle{1} \whitecircle{3} & $\checkmark$ & 07/2023 \\
            AgentSims~\cite{lin2023agentsims} & - & \whitecircle{2} & - & 08/2023 \\
            AgentBench~\cite{liu2023agentbench} & - & \whitecircle{3} & $\checkmark$ &  08/2023 \\
            BOLAA~\cite{liu2023bolaa} & - & \whitecircle{3} & $\checkmark$ &  08/2023 \\
            Gentopia~\cite{xu2023gentopia}& - & \whitecircle{3}  & $\checkmark$ &  08/2023 \\
            EmotionBench~\cite{huang2023emotionally}& \whitecircle{1} & - & $\checkmark$ &  08/2023 \\
            PTB~\cite{deng2023pentestgpt} & - & \whitecircle{4} & - & 08/2023 \\
                     \hline\hline
        \end{tabular}\label{evaluation}
}
\vspace{-0.cm}
\end{table*}

\subsection{Objective Evaluation}
Objective evaluation refers to assessing the capabilities of LLM-based autonomous agents using quantitative metrics that can be computed, compared and tracked over time. 
In contrast to subjective evaluation, objective metrics aim to provide concrete, measurable insights into the agent performance. 
For conducting objective evaluation, there are three important aspects, that is, the evaluation metrics, protocols and benchmarks.
In the following, we introduce these aspects more in detail.

\textbf{Metrics}: 
In order to objectively evaluate the effectiveness of the agents, designing proper metrics is significant, which may influence the evaluation accuracy and comprehensiveness.
Ideal evaluation metrics should precisely reflect the quality of the agents, and align with the human feelings when using them in real-world scenarios.
In existing work, we can conclude the following representative evaluation metrics.
(1) \textit{Task success metrics:} These metrics measure how well an agent can complete tasks and achieve goals. Common metrics include success rate~\cite{zhang2023building,yao2022react,shinn2023reflexion,liu2023llmp+}, reward/score~\cite{zhang2023building,yao2022react,mehta2023improving}, coverage~\cite{zhu2023ghost}, and accuracy/error rate~\cite{park2022social,qian2023communicative,aher2023using,hu2023chatdb}. 
Depending on the scenario, accuracy may reflect aspects such as program executability~\cite{qian2023communicative} or task validity~\cite{aher2023using}.
Higher values across these task success metrics indicate greater task completion ability.
(2) \textit{Human similarity metrics:} 
These metrics quantify the degree to which the agent behaviors closely resembles those of humans by emphasizing various aspects related to human traits, such as coherent~\cite{ziems2023can}, fluent~\cite{ziems2023can}, dialogue similarities with human~\cite{park2022social} and human acceptance rate~\cite{aher2023using}. 
Higher similarity suggests better human simulation performance.
(3) \textit{Efficiency metrics:} 
In contrast to the aforementioned metrics used to evaluate the agent effectiveness, these metrics aim to assess the efficiency of agent. 
Commonly considered metrics encompass the cost associated with development~\cite{qian2023communicative} and training efficiency~\cite{zhu2023ghost,wang2023voyager}.

\textbf{Protocols}: 
In addition to the evaluation metrics, another important aspect for objective evaluation is how to leverage these metrics.
In the previous work, we can identify the following commonly used evaluation protocols:
(1) \textit{Real-world simulation:} 
In this method, the agents are evaluated within immersive environments like games and interactive simulators. 
The agents are required to perform tasks autonomously, and then metrics like task success rate and human similarity are leveraged to evaluate the capability of the agents based on their trajectories and completed objectives~\cite{zhang2023building,zhu2023ghost,yao2022react,wang2023voyager,mehta2023improving,wang2023describe,liang2023tachikuma,yao2022webshop,zhang2023mobileenv,shinn2023reflexion}.
By simulating real-world scenarios, this approach aims to provide a comprehensive evaluation of the agents’ practical capabilities.
(2) \textit{Social evaluation}: 
This method utilizes metrics to assess social intelligence based on the agent interactions in simulated societies. 
Various approaches have been adopted, such as collaborative tasks to evaluate teamwork skills, debates to analyze argumentative reasoning, and human studies to measure social aptitude~\cite{park2022social,aher2023using,choi2023socket,lin2023agentsims}. These approaches analyze qualities such as coherence, theory of mind, and social IQ to assess agents' capabilities in areas including cooperation, communication, empathy, and mimicking human social behavior. By subjecting agents to complex interactive settings, social evaluation provides valuable insights into agents' higher-level social cognition.
(3) \textit{Multi-task evaluation:} In this method, people use a set of diverse tasks from different domains to evaluate the agent, which can effectively measure the agent generalization capability in open-domain environments~\cite{argyle2023out,choi2023socket,qin2023bmtools,liu2023agentbench,liu2023bolaa,yao2022webshop,zhang2023mobileenv,shinn2023reflexion,chen2023two}.
(4) \textit{Software testing:} In this method, researchers evaluate the agents by letting them conduct tasks such as software testing tasks, such as generating test cases, reproducing bugs, debugging code, and interacting with developers and external tools~\cite{jalil2023chatgpt,kang2023large,feldt2023towards}. Then, one can use metrics like test coverage and bug detection rate to measure the effectiveness of LLM-based agents.

\textbf{Benchmarks}: 
Given the metrics and protocols, a crucial aspect of evaluation is the selection of appropriate benchmarks. Over time, various benchmarks have been introduced to assess the capabilities of LLM-based agents across diverse domains and scenarios.
Many studies employ environments such as ALFWorld~\cite{yao2022react}, IGLU~\cite{mehta2023improving}, and Minecraft~\cite{zhu2023ghost,wang2023voyager,wang2023describe} to evaluate agent capabilities in interactive and task-oriented simulations.
Tachikuma~\cite{liang2023tachikuma} evaluates LLMs’ ability to infer and understand complex interactions involving multiple characters and novel objects through TRPG game logs.
AgentBench~\cite{liu2023agentbench} provides a comprehensive framework for evaluating LLMs as autonomous agents across diverse environments. It represents the first systematic assessment of LLMs as agents on real-world challenges across diverse domains. 
SocKET~\cite{choi2023socket} is a comprehensive benchmark for evaluating the social capabilities of LLMs across 58 tasks covering five categories of social information such as humor and sarcasm, emotions and feelings, credibility, etc.
AgentSims~\cite{lin2023agentsims} is a versatile framework for evaluating LLM-based agents, where one can flexibly design the agent planning, memory and action strategies, and measure the effectiveness of different agent modules in interactive environments.
ToolBench~\cite{qin2023bmtools} focuses on assessing and enhancing language models' ability to use tools, featuring 16,464 real-world RESTful APIs and diverse instructions tailored for single- and multi-tool scenarios.
WebShop~\cite{yao2022webshop} develops a benchmark for evaluating LLM-based agents in terms of their capabilities for product search and retrieval, which is constructed using a collection of 1.18 million real-world items.
Mobile-Env~\cite{zhang2023mobileenv} serves as an extendable interactive platform designed to evaluate the multi-step interaction capabilities of LLM-based agents.
WebArena~\cite{zhou2023webarena} offers a comprehensive website environment that spans multiple domains. Its purpose is to evaluate agents in an end-to-end fashion and determine the accuracy of their completed tasks.
GentBench~\cite{xu2023gentopia} is crafted to evaluate the agent capabilities, including their reasoning, safety, and efficiency, when utilizing tools to complete complex tasks.
RocoBench~\cite{mandi2023roco} comprises six tasks that evaluate multi-agent collaboration across diverse scenarios, emphasizing communication and coordination strategies to assess adaptability and generalization in cooperative robotics.
EmotionBench~\cite{huang2023emotionally} evaluates the emotion appraisal ability of LLMs, i.e., how their feelings change when presented with specific situations. It collects over 400 situations that elicit eight negative emotions and measures the emotional states of LLMs and human subjects using self-report scales.
PEB~\cite{deng2023pentestgpt} is tailored for assessing LLM-based agents in penetration testing scenarios, comprising 13 diverse targets from leading platforms. It offers a structured evaluation across varying difficulty levels, reflecting real-world challenges for agents.
ClemBench~\cite{chalamalasetti2023clembench} contains five Dialogue Games to assess LLMs' ability as a player.
E2E~\cite{banerjee2023benchmarking} serves as an end-to-end benchmark for testing the accuracy and usefulness of chatbots.
\begin{remark}
% Objective evaluation allows for the quantitative assessment of LLM-based agent capabilities using diverse metrics.
Objective evaluation facilitates the quantitative analysis of capabilities in LLM-based agents through a variety of metrics.
While current techniques can not perfectly measure all types of agent capabilities, objective evaluation provides essential insights that complement subjective assessment.
% The ongoing progress in objective evaluation benchmarks and methodology will further advance the development and understanding of LLM-based autonomous agents.
Continued advancements in benchmarks and methodologies for objective evaluation will enhance the development and understanding of LLM-based autonomous agents further.

In the above sections, we introduce both subjective and objective strategies for LLM-based autonomous agent evaluation. 
The evaluation of the agents play significant roles in this domain. However, both subjective and objective evaluation have their own strengths and weakness. 
Maybe, in practice, they should be combined to comprehensively evaluate the agents. 
We summarize the correspondence between the previous work and these evaluation strategies in Table~\ref{evaluation}.
\end{remark}

\vspace{-0.2cm}
\section{Related Surveys}\label{rel}

With the vigorous development of large language models, a variety of comprehensive surveys have emerged, providing detailed insights into various aspects. ~\cite{zhao2023survey} extensively introduces the background, main findings, and mainstream technologies of LLMs, encompassing a vast array of existing works. On the other hand, ~\cite{yang2023harnessing} primarily focus on the applications of LLMs in various downstream tasks and the challenges associated with their deployment. Aligning LLMs with human intelligence is an active area of research to address concerns such as biases and illusions. ~\cite{wang2023aligning} have compiled existing techniques for human alignment, including data collection and model training methodologies. Reasoning is a crucial aspect of intelligence, influencing decision-making, problem-solving, and other cognitive abilities. 
~\cite{huang2023reasoning} presents the current state of research on LLMs' reasoning abilities, exploring approaches to improve and evaluate their reasoning skills. ~\cite{mialon2023augmented} propose that language models can be enhanced with reasoning capabilities and the ability to utilize tools, termed Augmented Language Models (ALMs). They conduct a comprehensive review of the latest advancements in ALMs. As the utilization of large-scale models becomes more prevalent, evaluating their performance is increasingly critical. 
~\cite{chang2023survey} shed light on evaluating LLMs, addressing what to evaluate, where to evaluate, and how to assess their performance in downstream tasks and societal impact. 
~\cite{chang2024language} also discusses the capabilities and limitations of LLMs in various downstream tasks.
The aforementioned research encompasses various aspects of large models, including training, application, and evaluation. However, prior to this paper, no work has specifically focused on the rapidly emerging and highly promising field of LLM-based Agents. In this study, we have compiled 100 relevant works on LLM-based Agents, covering their construction, applications, and evaluation processes.

\section{Challenges}\label{disscusion}
While previous work on LLM-based autonomous agent has obtained many remarkable successes, this field is still at its initial stage, and there are several significant challenges that need to be addressed in its development. In the following, we present many representative challenges.

\subsection{Role-playing Capability}
Different from traditional LLMs, autonomous agent usually has to play as specific roles (\emph{e.g.}, program coder, researcher and chemist) for accomplishing different tasks.
Thus, the capability of the agent for role-playing is very important. 
Although LLMs can effectively simulate many common roles such as movie reviewers, there are still various roles and aspects that they struggle to capture accurately.
To begin with, LLMs are usually trained based on web-corpus, thus for the roles which are seldom discussed on the web or the newly emerging roles, LLMs may not simulate them well. 
In addition, previous research~\cite{fischer2023reflective} has shown that existing LLMs may not well model the human cognitive psychology characters, leading to the lack of self-awareness in conversation scenarios.
Potential solution to these problems may include fine-tuning LLMs or carefully designing the agent prompts/architectures~\cite{li2023emotionprompt}.
For example, one can firstly collect real-human data for uncommon roles or psychology characters, and then leverage it to fine-tune LLMs.
However, how to ensure that fine-tuned model still perform well for the common roles may pose further challenges.
Beyond fine-tuning, one can also design tailored agent prompts/architectures to enhance the capability of LLM on role-playing.
However, finding the optimal prompts/architectures is not easy, since their designing spaces are too large.

\vspace{-0.1cm}
\subsection{Generalized Human Alignment}
% \cite{deshpande2023toxicity} may be relevant 
Human alignment has been discussed a lot for traditional LLMs.
In the field of LLM-based autonomous agent, especially when the agents are leveraged for simulation, we believe this concept should be discussed more in depth.
In order to better serve human-beings, traditional LLMs are usually fine-tuned to be aligned with correct human values, for example, the agent should not plan to make a bomb for avenging society.
However, when the agents are leveraged for real-world simulation, an ideal simulator should be able to honestly depict diverse human traits, including the ones with incorrect values. 
Actually, simulating the human negative aspects can be even more important, since an important goal of simulation is to discover and solve problems, and without negative aspects means no problem to be solved.
For example, to simulate the real-world society, we may have to allow the agent to plan for making a bomb, and observe how it will act to implement the plan as well as the influence of its behaviors.
Based on these observations, people can make better actions to stop similar behaviors in real-world society.
Inspired by the above case, maybe an important problem for agent-based simulation is how to conduct generalized human alignment, that is, for different purposes and applications, the agent should be able to align with diverse human values.
However, existing powerful LLMs including ChatGPT and GPT-4 are mostly aligned with unified human values.
Thus, an interesting direction is how to ``realign'' these models by designing proper prompting strategies.

\vspace{-0.1cm}
\subsection{Prompt Robustness}
To ensure rational behavior in agents, it's a common practice for designers to embed supplementary modules, such as memory and planning modules, into LLMs. However, the inclusion of these modules necessitates the development of more complex prompts in order to facilitate consistent operation and effective communication. Previous research~\cite{zhuo2023robustness,gekhman2023robustness} has highlighted the lack of robustness in prompts for LLMs, as even minor alterations can yield substantially different outcomes. This issue becomes more pronounced when constructing autonomous agents, as they encompass not a single prompt but a prompt framework that considers all modules, wherein the prompt for one module has the potential to influence others.
Moreover, the prompt frameworks can vary significantly across different LLMs. 
% Developing a unified and robust prompt framework that can be applied to various LLMs is an important yet unresolved issue.
The development of a unified and resilient prompt framework applicable across diverse LLMs remains a critical and unresolved challenge.
There are two potential solutions to the aforementioned problems: (1) manually crafting the essential prompt elements through trial and error, or (2) automatically generating prompts using GPT.

\vspace{-0.1cm}
\subsection{Hallucination}
Hallucination poses a fundamental challenge for LLMs, characterized by the models' tendency to produce false information with a high level of confidence. This challenge is not limited to LLMs alone but is also a significant concern in the domain of autonomous agents. For instance, in~\cite{ji2023survey}, it was observed that when confronted with simplistic instructions during code generation tasks, the agent may exhibit hallucinatory behavior. Hallucination can lead to serious consequences such as incorrect or misleading code, security risks, and ethical issues \cite{ji2023survey}. To mitigate this issue, incorporating human correction feedback directly into the iterative process of human-agent interaction presents a viable approach~\cite{hong2023metagpt}.
More discussions on the hallucination problem can be seen in~\cite{zhao2023survey}.

\vspace{-0.1cm}
\subsection{Knowledge Boundary}
A pivotal application of LLM-based autonomous agents lies in simulating diverse real-world human behaviors~\cite{park2023generative}.
The study of human simulation has a long history, and the recent surge in interest can be attributed to the remarkable advancements made by LLMs, which have demonstrated significant capabilities in simulating human behavior.
However, it is important to recognize that the power of LLMs may not always be advantageous. Specifically, an ideal simulation should accurately replicate human knowledge. In this context, LLMs may display overwhelming capabilities, being trained on a vast corpus of web knowledge that far exceeds what an average individual might know.
The immense capabilities of LLMs can significantly impact the effectiveness of simulations. For instance, when attempting to simulate user selection behaviors for various movies, it is crucial to ensure that LLMs assume a position of having no prior knowledge of these movies. However, there is a possibility that LLMs have already acquired information about these movies. Without implementing appropriate strategies, LLMs may make decisions based on their extensive knowledge, even though real-world users would not have access to the contents of these movies beforehand.
Based on the above example, we may conclude that for building believable agent simulation environment, an important problem is how to constrain the utilization of user-unknown knowledge of LLM.

\vspace{-0.cm}
\subsection{Efficiency}
Due to their autoregressive architecture, LLMs typically have slow inference speeds.
% In practical applications, agents may require multiple queries to LLMs for each discrete action, involving processes such as extracting information from memory modules, formulating plans before executing actions, and more.
However, the agent may need to query LLMs for each action multiple times, such as extracting information from memory, make plans before taking actions and so on.
Consequently, the efficiency of agent actions is greatly affected by the speed of LLM inference.
% Deploying multiple agents with the same API key can further significantly increase the time cost.
% As a result, the speed of LLM inference significantly influences the efficiency of agent actions. Additionally, deploying several agents with the same API key can notably escalate the time cost.

\section{Conclusion}\label{conclu}
In this survey, we systematically summarize existing research in the field of LLM-based autonomous agents.
We present and review these studies from three aspects 
including the construction, application, and evaluation of the agents.
For each of these aspects, we provide a detailed taxonomy to draw connections among the existing research, summarizing the major techniques and their development histories.
In addition to reviewing the previous work, we also propose several challenges in this field, which are expected to guide potential future directions.

\section*{Acknowledgement}
This work is supported in part by National Natural Science Foundation of China (No. 62102420), Beijing Outstanding Young Scientist Program NO. BJJWZYJH012019100020098, Intelligent Social Governance Platform, Major Innovation \& Planning Interdisciplinary Platform for the "Double-First Class" Initiative, Renmin University of China, Public Computing Cloud, Renmin University of China, fund for building world-class universities (disciplines) of Renmin University of China, Intelligent Social Governance Platform.

\bibliographystyle{fcs}
\bibliography{ref}

\begin{biography}{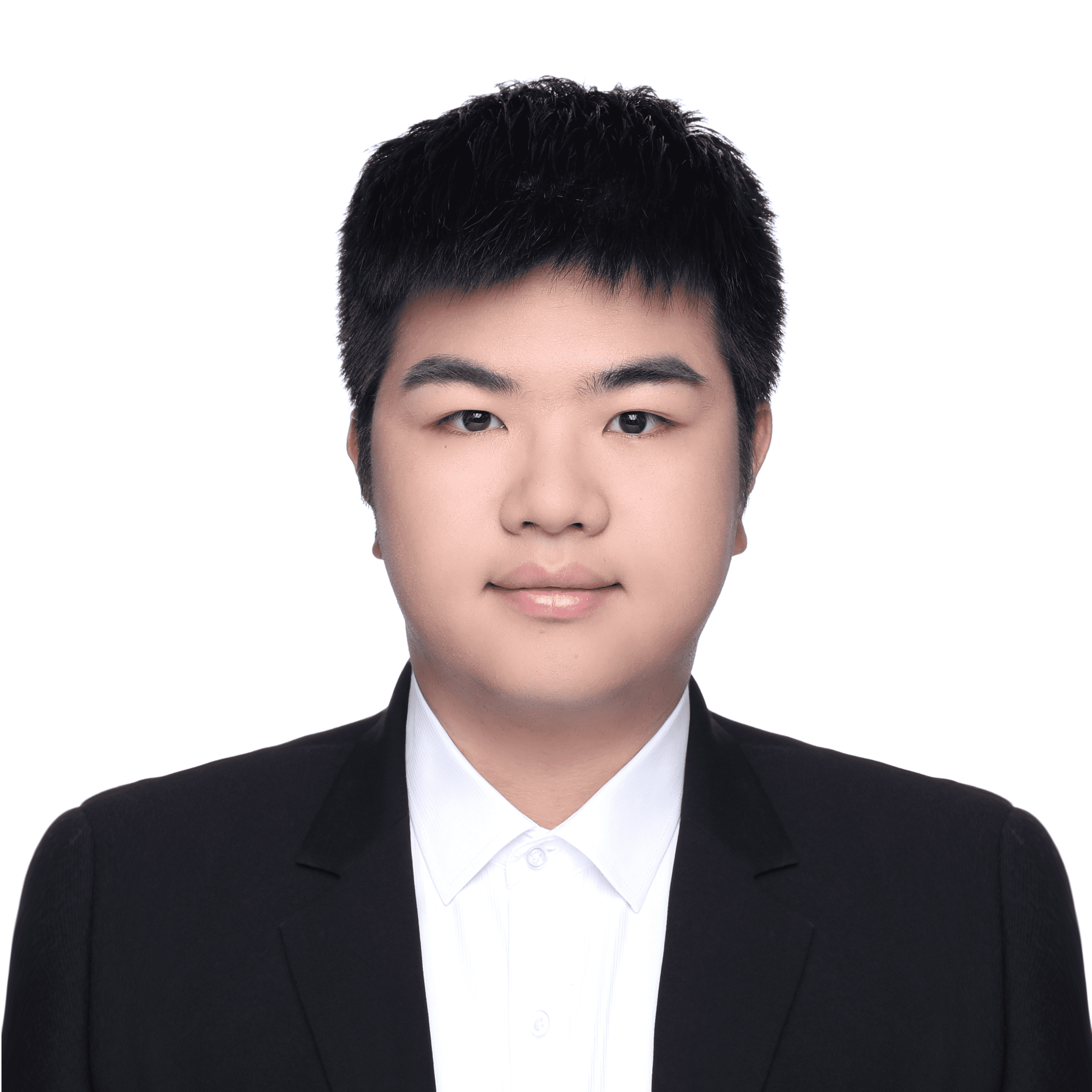}
Lei Wang is a Ph.D. candidate at Renmin University of China, Beijing. His research focuses on recommender systems and agent-based large language models.
\end{biography}

\begin{biography}{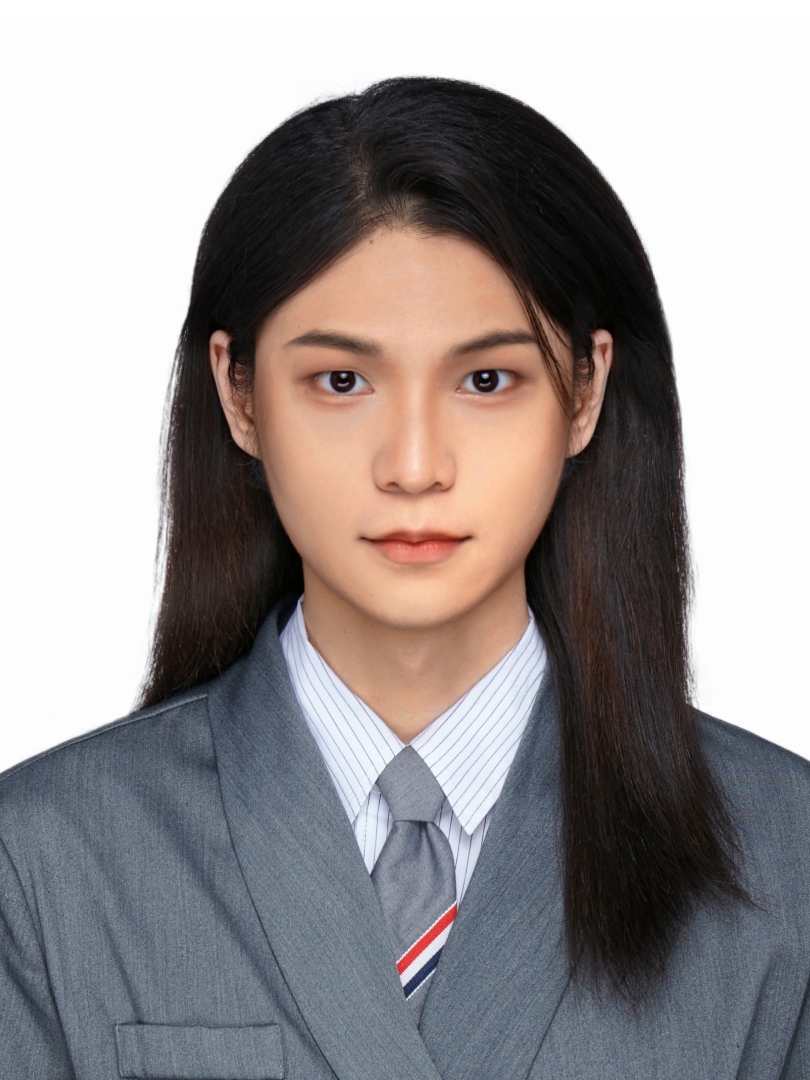}
{Chen Ma}
is currently pursuing a Master's degree at Renmin University of
    China, Beijing, China. His research interests include recommender system, agent based on large language model.
\end{biography}
\begin{biography}{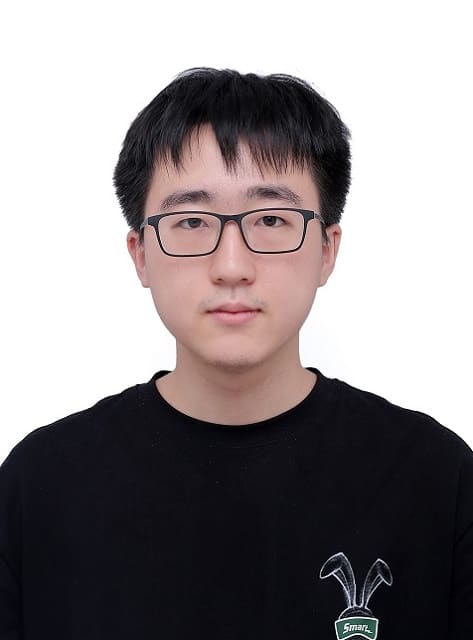}
{Xueyang Feng}
is currently studying for a Ph.D. degree at Renmin University of
    China, Beijing, China. His research interests include recommender system, agent based on large language model.
\end{biography}
\begin{biography}{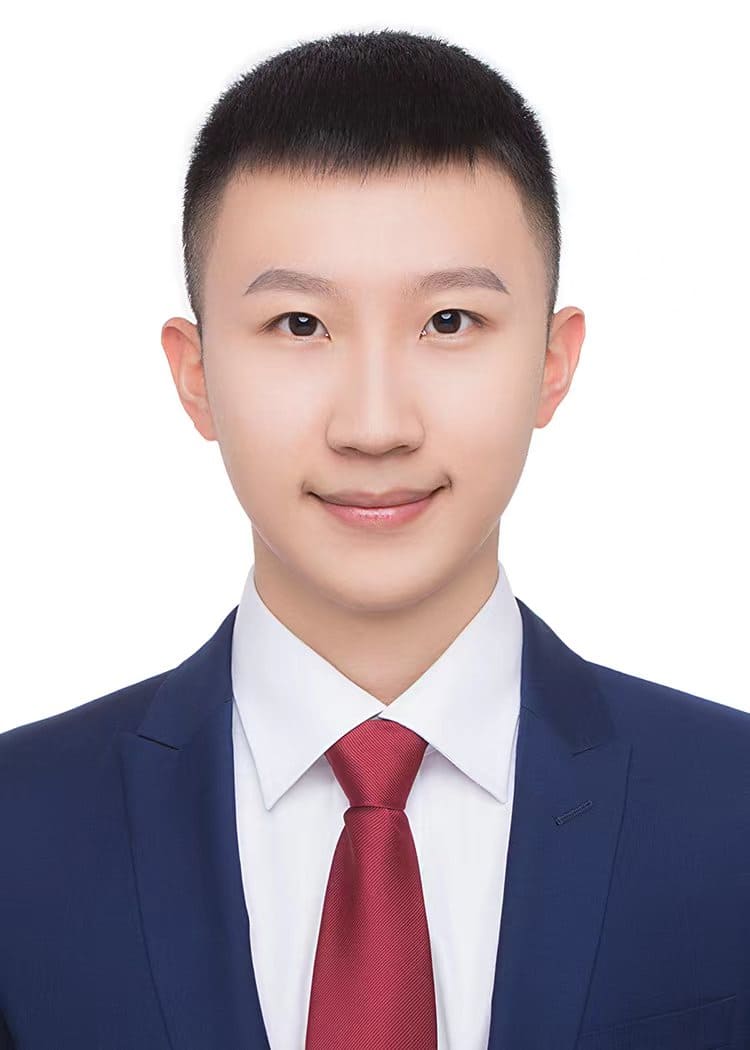}
{Zeyu Zhang}
is currently pursuing a Master's degree at Renmin University of
    China, Beijing, China. His research interests include recommender system, causal inference, agent based on large language model.
\end{biography}
\begin{biography}{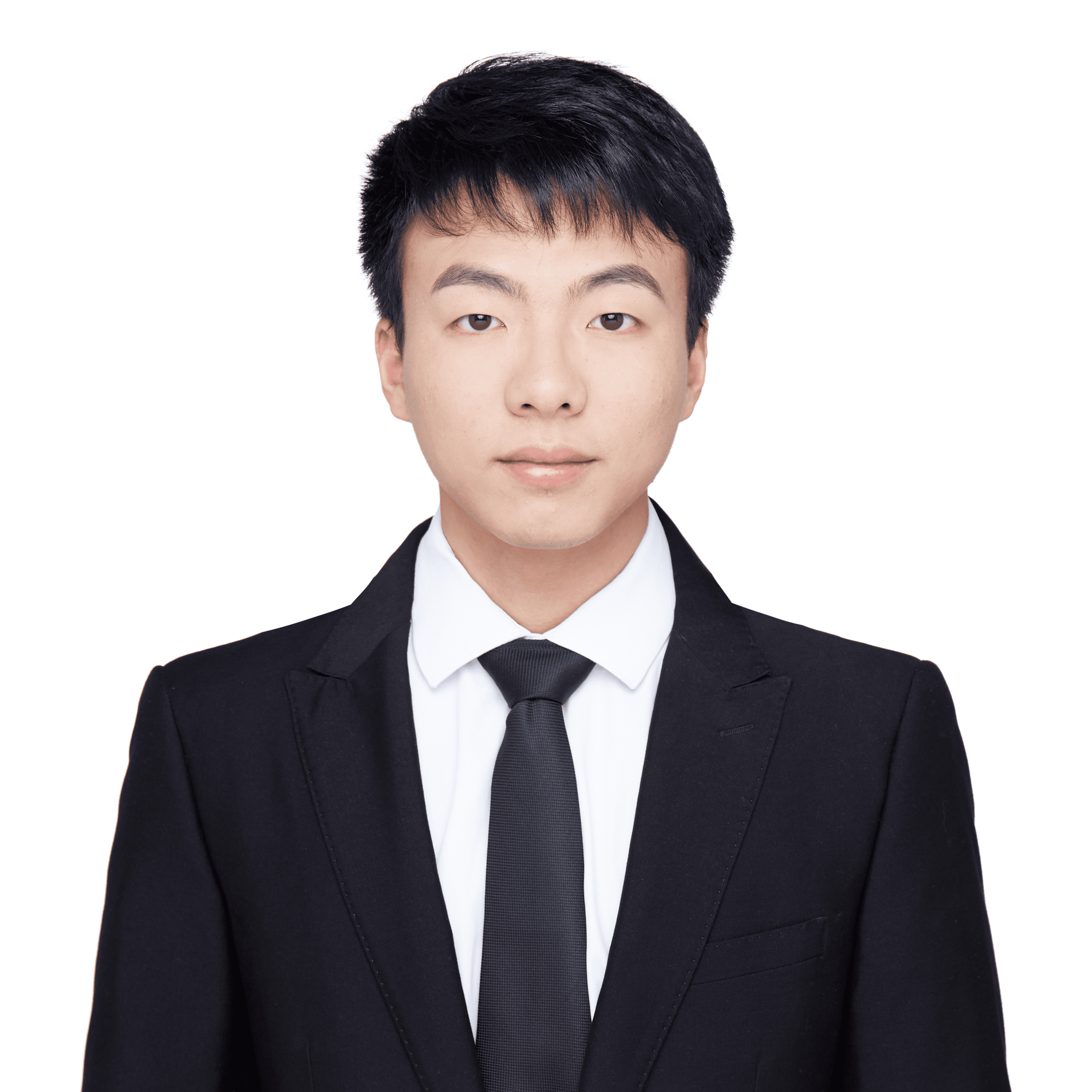}
{Hao Yang}
is currently studying for a Ph.D. degree at Renmin University of
    China, Beijing, China. His research interests include recommender system, causal inference.
\end{biography}
\begin{biography}{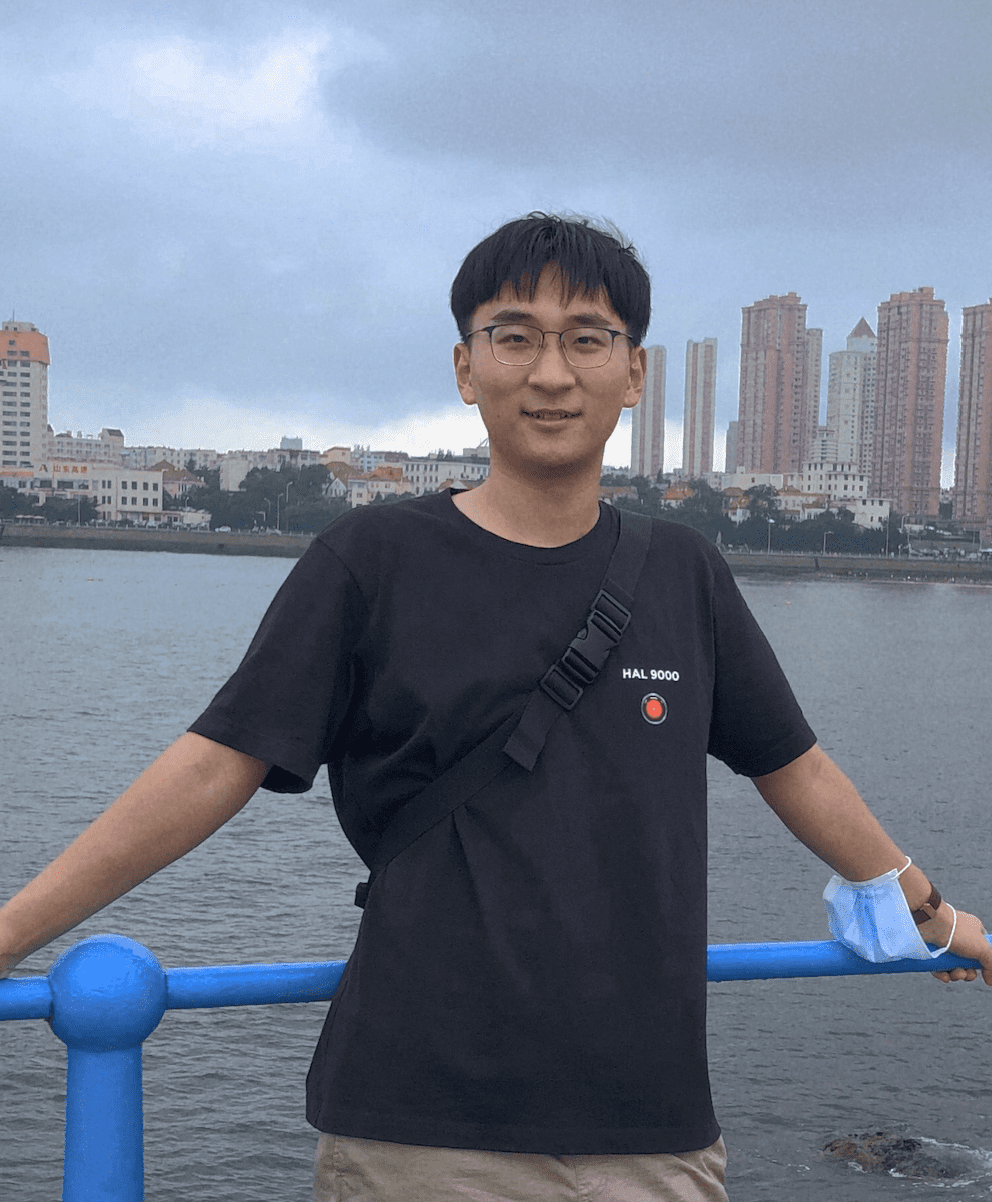}
{Jingsen Zhang}
is currently studying for a Ph.D. degree at Renmin University of
    China, Beijing, China. His research interests include recommender system.
\end{biography}
\begin{biography}{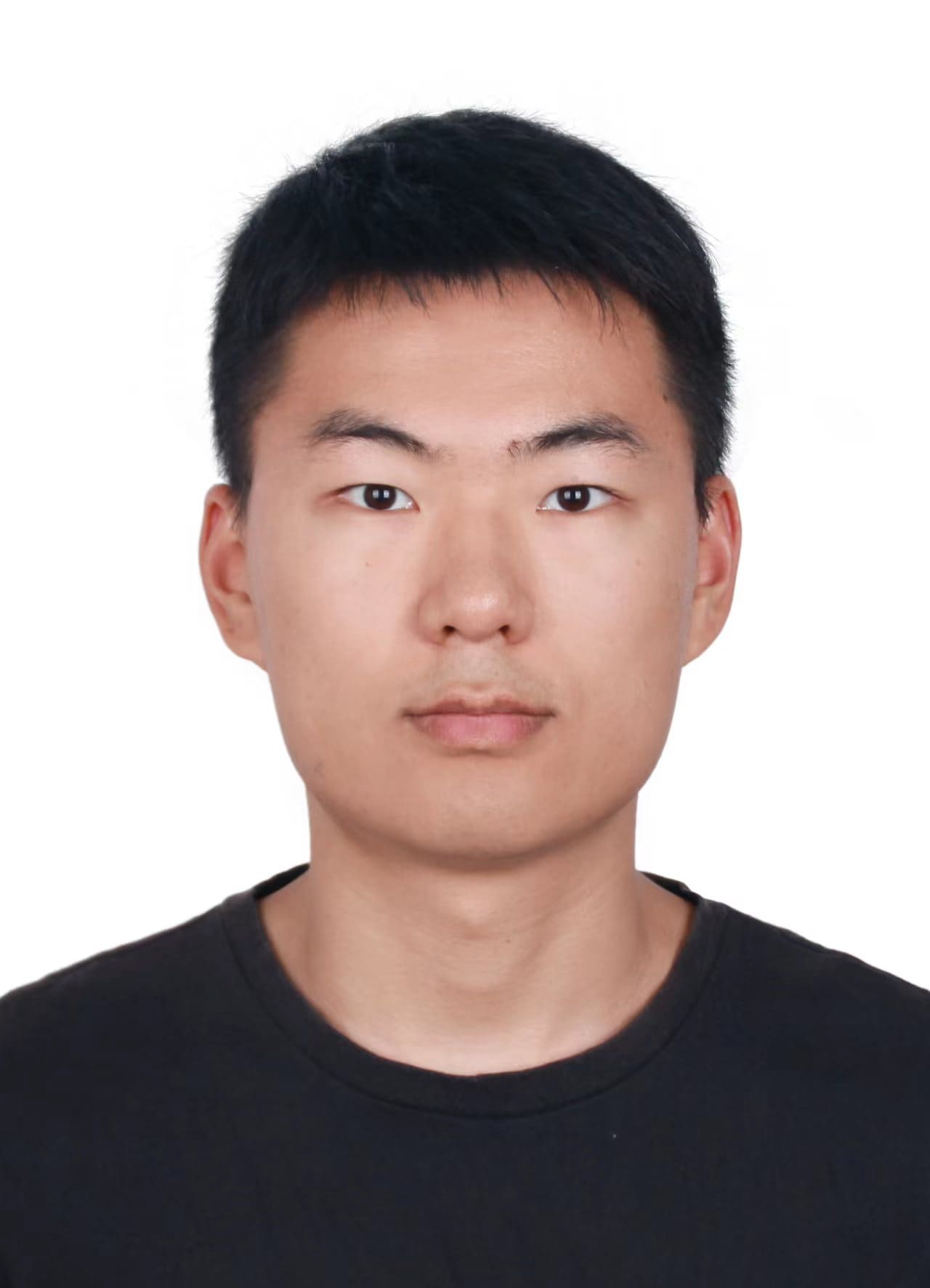}
{Zhi-Yuan Chen}
is pursuing his Ph.D. in Gaoling school of Artificial Intelligence, Renmin University of China. His research mainly focuses on language model reasoning and agent based on large language model.
\end{biography}
\begin{biography}{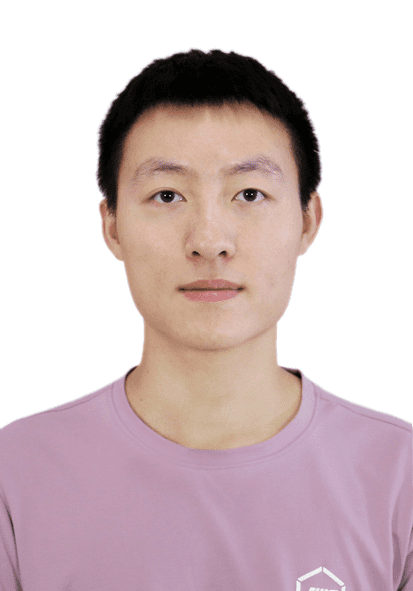}
{Jiakai Tang}
is currently pursuing a Master's degree at Renmin University of
    China, Beijing, China. His research interests include recommender system.
\end{biography}

\begin{biography}{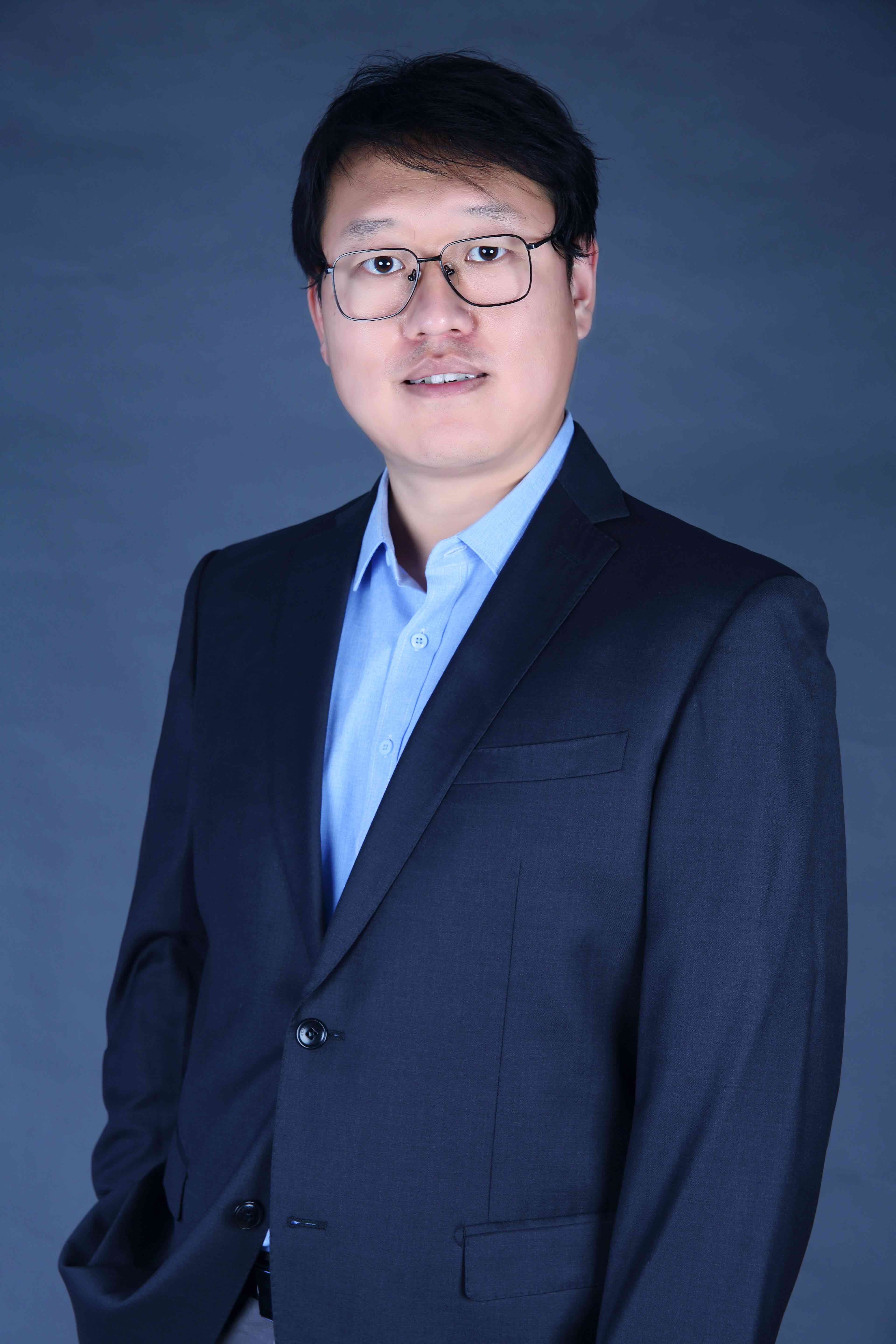}
{Xu Chen} obtained his PhD degree from Tsinghua University, China. Before joining Renmin University of China, he was a postdoc researcher at University College London, UK. In the period from March to September of 2017, he was studying at Georgia Institute of Technology, USA, as a visiting scholar. His research mainly focuses on the recommender system, reinforcement learning and causal inference.
\end{biography}

\begin{biography}{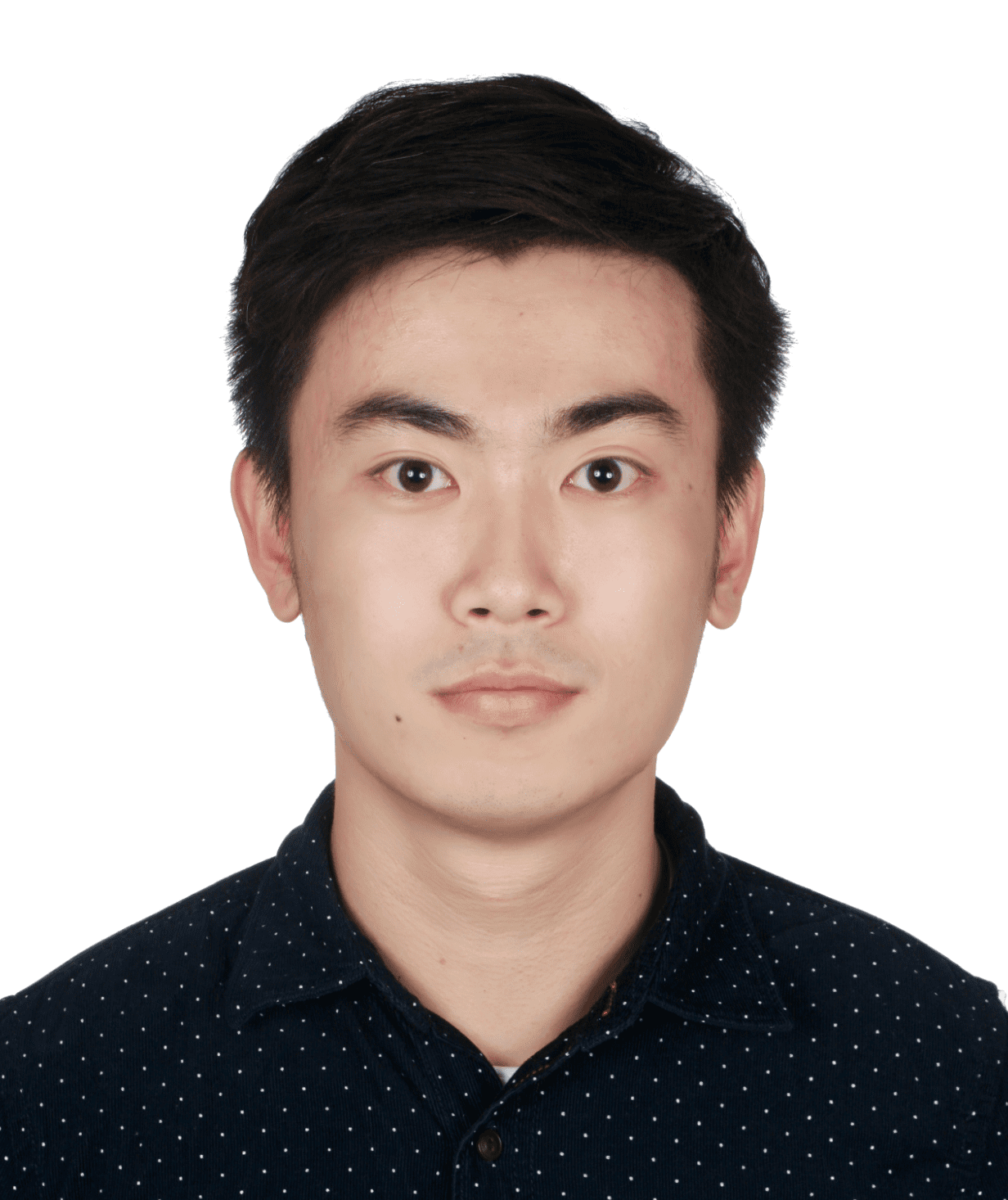}
{Yankai Lin} received his B.E. and Ph.D. degrees from Tsinghua University in 2014 and 2019. After that, he worked as a senior researcher in Tencent WeChat, and joined Renmin University of China in 2022 as a tenure-track assistant professor. His main research interests are pretrained models and natural language processing.
\end{biography}

\begin{biography}{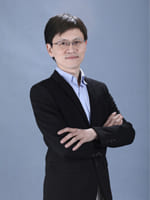}
{Wayne Xin Zhao} received his Ph.D. in Computer Science from Peking University in 2014. His research interests include data mining, natural language processing and information retrieval in general. The main goal is to study how to organize, analyze and mine user generated data for improving the service of real-world applications.
\end{biography}

\begin{biography}{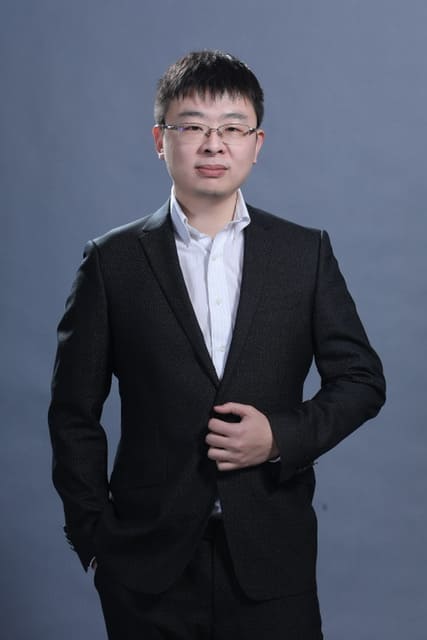}
{Zhewei Wei} received his Ph.D. of Computer Science and Engineering from Hong Kong University of Science and Technology. He did postdoctoral research in Aarhus University from 2012 to 2014, and joined Renmin University of China in 2014.
\end{biography}

\begin{biography}{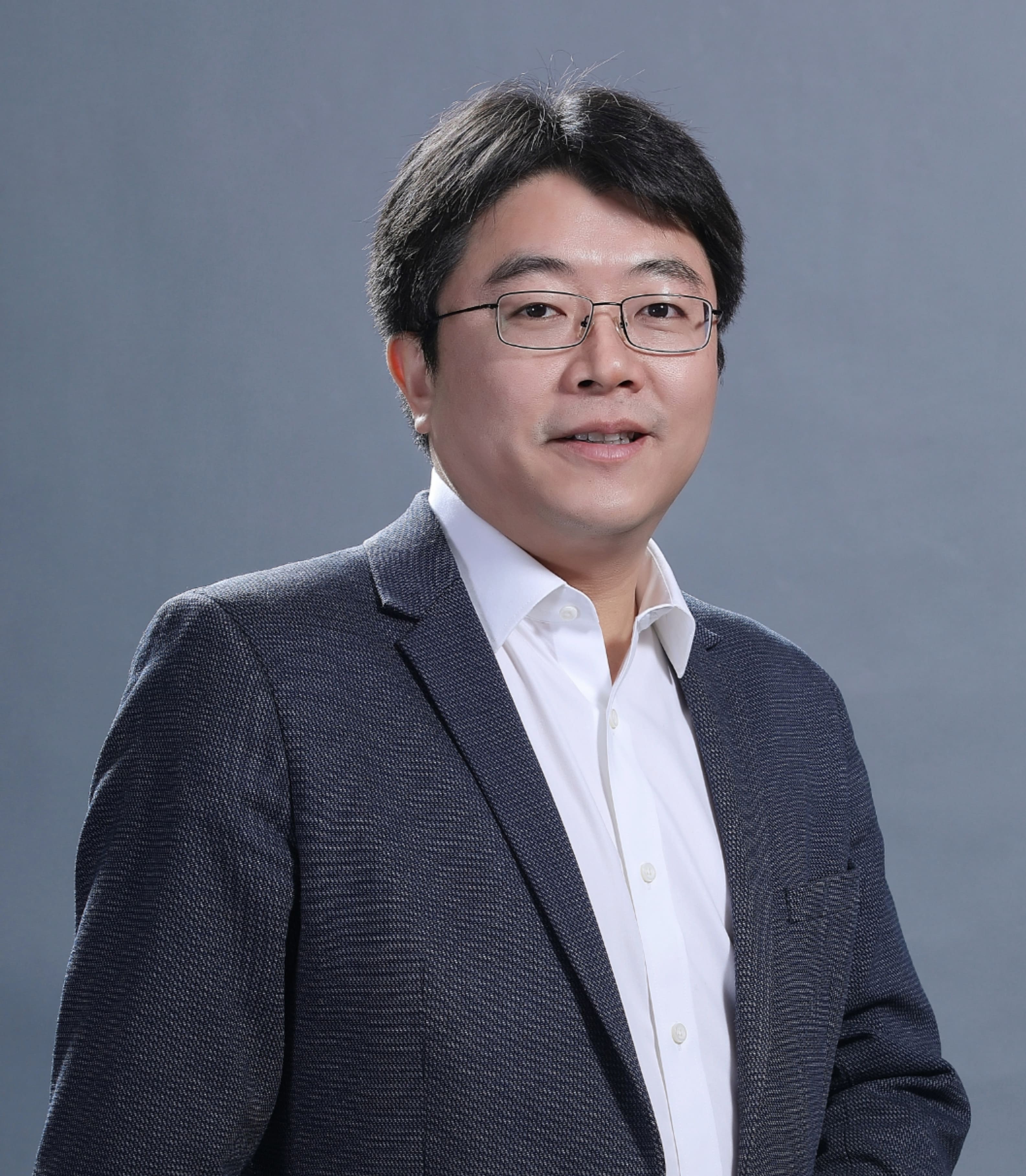}
{Ji-Rong Wen} is a full professor, the executive dean of Gaoling School of Artificial Intelligence, and the dean of School of Information at Renmin University of China. He has been working in the big data and AI areas for many years, and publishing extensively on prestigious international conferences and journals.
\end{biography}

\end{document}